%% file: example_paper.tex
\documentclass{article}

\usepackage{microtype}
\usepackage{graphicx}
\usepackage{caption}   
\usepackage{subcaption}
\usepackage{multirow}
\usepackage{booktabs} 
\usepackage{verbatim}
\usepackage{tikz}
\usepackage{makecell}
\usepackage{arydshln}
\usepackage{algorithm}
\usepackage[textsize=tiny]{todonotes}

\usepackage{hyperref}

\usepackage[accepted]{icml2025}

\usepackage{amsmath}
\usepackage{amssymb}
\usepackage{mathtools}
\usepackage{amsthm}

\usepackage[capitalize,noabbrev]{cleveref}

\theoremstyle{plain}

\theoremstyle{definition}

\theoremstyle{remark}

\icmltitlerunning{Beyond and Free from Diffusion: Invertible Guided Consistency Training}

\begin{document}

\twocolumn[
\icmltitle{Beyond and Free from Diffusion: Invertible Guided Consistency Training}




\begin{icmlauthorlist}
\icmlauthor{Chia-Hong Hsu}{brown}
\icmlauthor{Shiu-hong Kao}{hkust}
\icmlauthor{Randall Balestriero}{brown}

\end{icmlauthorlist}

\icmlaffiliation{brown}{Brown University, RI, US}
\icmlaffiliation{hkust}{HKUST, Hong Kong}

\icmlcorrespondingauthor{Chia-Hong Hsu}{chia\_hong\_hsu@brown.edu}
\icmlcorrespondingauthor{Shiu-hong Kao}{skao@cse.ust.hk}
\icmlcorrespondingauthor{Randall Balestriero}{randall\_balestriero@brown.edu}

\icmlkeywords{Consistency Models, Diffusion Models, Guidance, Image Editing, Consistency Training, ICML}

\vskip 0.3in

]




\begin{figure*}[t] 
    \centering
    \includegraphics[width=\textwidth]{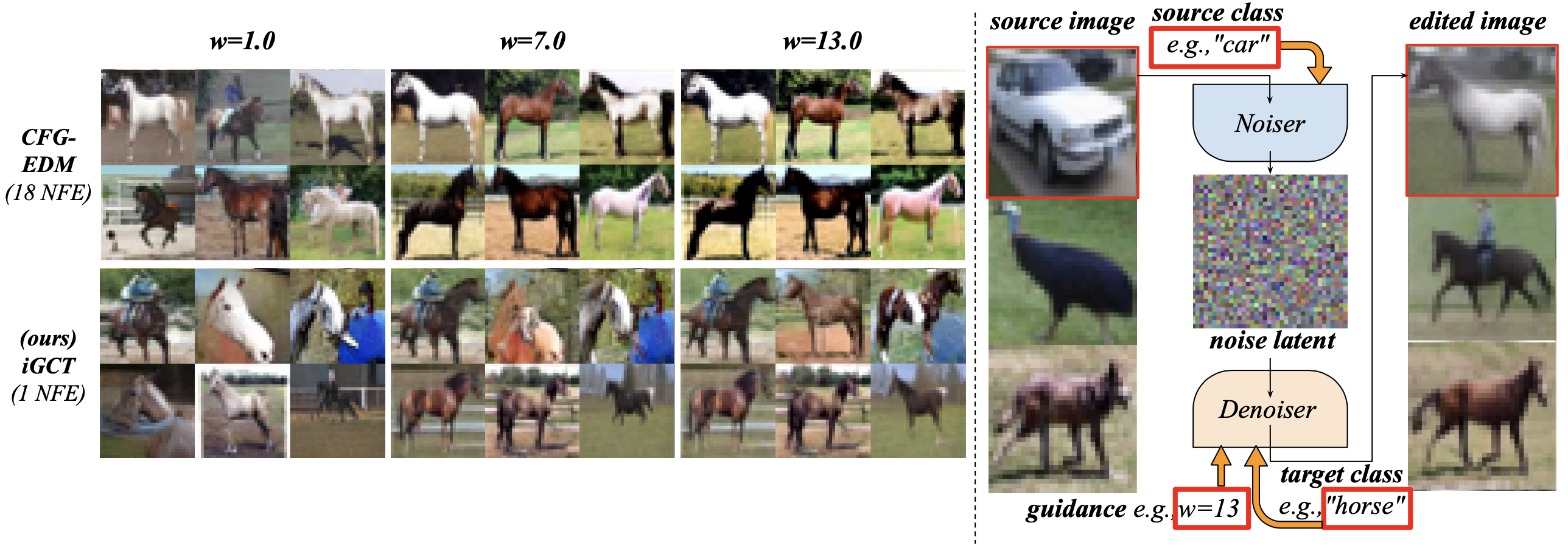}
    \caption{
        Guidance of EDM trained via Classifier-free Guidance (CFG) (top-left), and our \textit{Invertible Guided Consistency Training} (iGCT) (bottom-left). iGCT enables fast inverse-based image editing while preserving the source semantics (right). Unlike CFG, iGCT \textbf{eliminates the need for two-stage training} and removes contrast artifacts, achieving \textbf{better precision} and \textbf{FID} under high guidance (Fig. \ref{fig:results_fid_prec_rec}).
    }
    \vspace{-1.5em}
    \label{fig:teaser_horse_explain_v2}
\end{figure*}

\input{sec/0_abstract}    
\input{sec/1_intro}

\input{sec/2_background}
\input{sec/3_method}
\input{sec/4_results}

\input{sec/5_conclusion}

\newpage




\section*{Impact Statement}

The societal implications of this work are largely positive, as it contributes to creative industries, education, and research. However, as with any image generation technology, there is a risk of misuse, such as the creation of misleading or harmful content. We encourage the responsible use of this technology and emphasize the importance of ethical considerations in its deployment.

In summary, while our work primarily aims to advance the field of machine learning, we acknowledge the broader societal implications and encourage ongoing dialogue about the ethical use of image generation technologies.

\bibliography{example_paper}
\bibliographystyle{icml2025}

\newpage
\appendix
\onecolumn
\input{sec/6_appendix}
\end{document}

%% file: sec/0_abstract.tex
\begin{abstract}

Guidance in image generation steers models towards higher-quality or more targeted outputs, typically achieved in Diffusion Models (DMs) via Classifier-free Guidance (CFG). However, recent Consistency Models (CMs), which offer fewer function evaluations, rely on distilling CFG knowledge from pretrained DMs to achieve guidance, making them costly and inflexible. In this work, we propose \textbf{\textit{invertible Guided Consistency Training (iGCT)}}, a novel training framework for guided CMs that is entirely data-driven. iGCT, as a pioneering work, contributes to fast and guided image generation and editing without requiring the training and distillation of DMs, greatly reducing the overall compute requirements. iGCT addresses the saturation artifacts seen in CFG under high guidance scales. Our extensive experiments on CIFAR-10 \cite{article} and ImageNet64 \cite{chrabaszcz2017downsampledvariantimagenetalternative} show that iGCT significantly improves FID and precision compared to CFG. At a guidance of 13, iGCT improves precision to 0.8, while DM's drops to 0.47. Our work takes the first step toward enabling guidance and inversion for CMs without relying on DMs.
\end{abstract}

%% file: sec/1_intro.tex
\vspace{-0.5cm}
\section{Introduction}
\label{sec:intro}
\vspace{-0.15cm}

Diffusion Models (DMs) have emerged as a powerful class of generative models, demonstrating state-of-the-art performance in tasks such as image synthesis \cite{ho2020denoisingdiffusionprobabilisticmodels,song2020generativemodelingestimatinggradients,song2021scorebasedgenerativemodelingstochastic,song2022denoisingdiffusionimplicitmodels,rombach2022high,podell2023sdxlimprovinglatentdiffusion,lin2023magic3dhighresolutiontextto3dcontent} conditional generation \cite{Avrahami_2022,meng2022sdeditguidedimagesynthesis,mou2023t2iadapterlearningadaptersdig,bartal2023multidiffusionfusingdiffusionpaths,bashkirova2023masksketchunpairedstructureguidedmasked,huang2023composercreativecontrollableimage,zhang2023addingconditionalcontroltexttoimage,ni2023conditional,podell2023sdxlimprovinglatentdiffusion}, and precise image editing \cite{mokady2023null,miyake2023negative,wallace2023edict,han2024proxedit,garibi2024renoise,huberman2024edit}. Unlike GANs, which rely on adversarial training to generate samples \cite{goodfellow2014generativeadversarialnetworks,metz2016unrolled,gulrajani2017improved,arjovsky2017towards,karras2018progressivegrowinggansimproved,brock2019largescalegantraining,karras2019stylebasedgeneratorarchitecturegenerative,karras2020traininggenerativeadversarialnetworks,Karras2021}, diffusion models (DMs) use a stochastic process that incrementally adds Gaussian noise to data, then reverses this process to generate new samples. This iterative approach, while computationally intensive, enables greater control over sample quality and diversity through guided sampling methods \cite{ho2022classifier,dieleman2022guidance,ho2022imagenvideohighdefinition,Meng_2023_CVPR,karras2024guiding}. In contrast, GANs often suffer from mode collapse \cite{gulrajani2017improved,arjovsky2017towards,metz2016unrolled}, which hinders output diversity and training stability. Diffusion-based algorithms can thus offer a key advantage by balancing quality-diversity tradeoffs, frequently achieved with Classifier-free Guidance (CFG) \cite{ho2022classifier}. Both the role of diffusion noise scheduling and the network's convergence properties are essential for improving performance and applicability \cite{karras2022elucidating,lu2022dpmsolverfastodesolver,lu2023dpmsolverfastsolverguided,zheng2023dpm,karras2024analyzingimprovingtrainingdynamics,salimans2022progressivedistillationfastsampling,park2024textitjumpstepsoptimizingsampling,karras2024guidingdiffusionmodelbad}. 

Consistency Models (CMs), designed to accelerate the diffusion process, provide a more efficient alternative by reducing the number of function evaluations (NFE) needed for generation \cite{song2023consistency,kim2023consistency,kim2024generalized,li2024bidirectional,heek2024multistep,lu2024simplifyingstabilizingscalingcontinuoustime,lee2025truncatedconsistencymodels}. These models achieve this by aligning data points along the Probability-flow (PF) ODE of the stochastic forward process, using either consistency distillation (CD) or consistency training (CT) \cite{song2023consistency}. While CD models have been widely applied in real-time image synthesis and fast editing via distillation \cite{luo2023latent,starodubcev2024invertible}, the development of CT models remain relatively unexplored. This is primarily because CT, as a fully self-supervised and data-driven approach, requires a well-crafted training curriculum to be trained successfully ~\cite{song2023improved,ect}.

\begin{figure}[t!]
    \centering
    \begin{tikzpicture}[scale=0.9, transform shape]
        \draw[thick, ->] (0, 0) -- (7.5, 0) node[right] {Year};

        \draw[fill=blue!30] (0.5, 0) circle (0.15) node[below=0.4cm] {\parbox{2cm}{\centering \textbf{2019} \\ \small\textit{$\epsilon$-pred \cite{ho2020denoisingdiffusionprobabilisticmodels}}}};

        \draw[fill=blue!30] (2.5, 0) circle (0.15) node[below=0.4cm] {\parbox{2cm}{\centering \textbf{2021} \\ \small\textit{VE, VP \cite{song2021scorebasedgenerativemodelingstochastic}, iDDPM \cite{nichol2021improveddenoisingdiffusionprobabilistic} }}};

        \draw[fill=blue!30] (4.5, 0) circle (0.15) node[below=0.4cm] {\parbox{2cm}{\centering \textbf{2022} \\ \small\textit{DDIM \cite{song2022denoisingdiffusionimplicitmodels}, DPM-Solver v1\cite{lu2022dpmsolverfastodesolver}, EDM \cite{karras2022elucidating}, v-pred \cite{salimans2022progressivedistillationfastsampling}}}};

        \draw[fill=blue!30] (6.5, 0) circle (0.15) node[below=0.4cm] {\parbox{2cm}{\centering \textbf{2023} \\ \small\textit{DPM-Solver++ \cite{lu2023dpmsolverfastsolverguided}, DPM-Solver v3 \cite{lu2023dpmsolverfastsolverguided} }}};

    \end{tikzpicture}
    
    \begin{tikzpicture}[scale=0.7, transform shape]
        \node[draw, circle, fill=blue!20, minimum size=1.2cm, text width=2cm, align=center] (A) at (0, 0) {\textbf{DM sch.} EDM\cite{karras2022elucidating}};

        \node[draw, circle, fill=blue!20, minimum size=1.2cm, text width=2cm, align=center] (B) at (3.5, 0) {\textbf{CD methods} CD\cite{song2023consistency}, CTM\cite{kim2023consistency}};

        \draw[->, thick] (B) -- (A);




    \end{tikzpicture}
    \vspace{-1em}
    \caption{Timeline of key developments in diffusion training/sampling schedulers (top). The schedulers used by pretrained DMs that CD depends on (bottom). The fields of DM and CT are advancing rapidly in multiple directions, and tightly coupling CM with DM would slow down progress. To train a CM under a particular diffusion scheduler, CD-based CMs rely on a teacher DM trained under the same setting, while \textbf{CT is free of such constraint}.} 
    \vspace{-1.5em}
    \label{fig:dm&cd}
\end{figure}
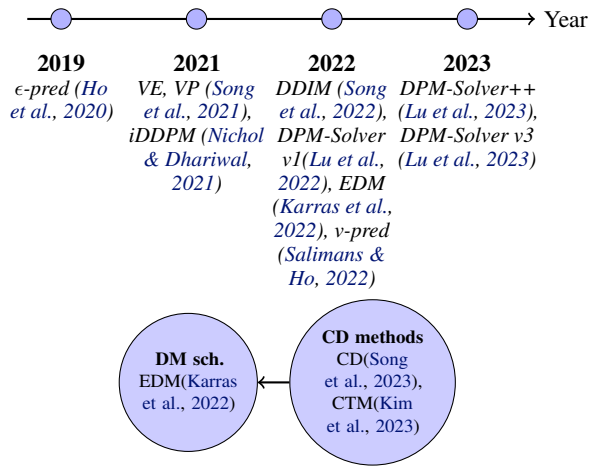

CT, however, offers distinct advantages. First, unlike CD, CT can be trained with any underlying diffusion schedulers. The design space for DM training, sampling, and preconditioning \cite{karras2022elucidating} is an ongoing area of research that has attracted researchers interest ever since the diffusion surge. Therefore, CT's flexibility to train with improved designs makes it more attractive than CD by dropping the dependence of a state-of-the-art DM. An improved diffusion scheduler can be directly utilized in CT, while CD depends on the existence of a DM (Fig.~\ref{fig:dm&cd}). Second, recent studies show that CT, when improved with advanced training techniques, outperforms CD in image generation on FID metrics \cite{song2023improved,ect}. This underscores its potential to surpass CD in both efficiency and quality, making CT as a promising alternative for advancing fast generative modeling. 

Our study fills the missing piece in establishing a comprehensive theory for CT, a data-driven approach for \textit{guidance learning} in 1-step/few-step image generation. We present \textbf{\textit{Invertible Guided Consistency Training}} (iGCT), a novel framework that enables CMs to perform guided generation without requiring DM distillation at training. In training, iGCT decouples the target clean sample from the source to capture effects of both unconditional and conditional noise. For invertibility, iGCT incorporates an independent component, the noiser, which maps images to noise in a single step. This same-dimensional noise, termed as \textit{noise latent} in our paper, serves as a deterministic representation of the input image. Removing the dependency on a teacher DM, iGCT provides greater flexibility and efficiency by eliminating the need of a two-stage training procedure. We show that guidance learned via iGCT removes the high-contrast artifacts (Figs.~\ref{fig:teaser_horse_explain_v2} and \ref{fig:1d_cfg_igct}), a well-known \textit{limitation} observed for CFG trained DMs \cite{ho2022imagenvideohighdefinition,saharia2022photorealistictexttoimagediffusionmodels,bradley2024classifierfreeguidancepredictorcorrector,kynkaanniemi2024applying,karras2024guiding}. Consequently, iGCT achieves higher FID's and Precision/Recall under guided generation compared to multi-step DMs and their distilled CM counterparts. In line with the parallel work on few-step image editing from iCD \cite{starodubcev2024invertible}, we demonstrate that iGCT is effective for 1-step image editing. Following common editing frameworks that applies high guidance after DDIM inversion \cite{mokady2023null,miyake2023negative,han2024proxedit,starodubcev2024invertible}, iGCT is able to match the target class semantics while preserving features inversed from the source image. To summarize, our contributions are as follows:
\begin{itemize}
    \vspace{-0.5em}
    \item We introduce \textbf{\textit{invertible Guided Consistency Training}} (iGCT), a DM independent approach that incorporates guidance into CMs (Sec. \ref{sec:method} and Appendix \ref{appendix:iGCT}). 
    \vspace{-0.5em}
    \item We demonstrate iGCT's ability to trade-off diversity for precision as guidance strength increases, beating CFG trained DMs and distilled CMs by eliminating saturated artifacts under high guidance scale (Figs. \ref{fig:results_guidance} and  \ref{fig:results_fid_prec_rec}).
    \vspace{-0.5em}
    \item We present iGCT’s invertibility, a disjoint contribution that enables efficient inversion and generation using class conditioning, offering a promising framework for real-time image editing (Sec. \ref{sec:image-editing}).
\end{itemize}

%% file: sec/2_background.tex
\vspace{-0.3cm}
\section{Background and Preliminaries}
\vspace{-0.15cm}
This section introduces the foundational concepts relevant to our methodology and baselines on CMs, CFG, and inversion-based image editing. In the remaining discussion, we denote \(\boldsymbol{x}_0\) as a sample from the target data distribution, denoted as \(p_\text{data}(\boldsymbol{x})\). 

\vspace{-0.15cm}
\subsection{EDM and Consistency Models}
\label{sec:edm_cm}
\vspace{-0.15cm}

In this paper, we mainly consider EDM \cite{karras2022elucidating}, a popular formulation of DM, diffusion scheduler, and ODE widely used by CMs \cite{song2023consistency,song2023improved,li2024bidirectional,ect}. We introduce two ways of learning discrete-time CMs, CD and CT, originally in the first CM paper \cite{song2023consistency}, as well as the continuous-time CM from ECT \cite{ect}.  

\noindent{\bf EDM.} The stochastic forward process of EDM is described by, \(\text{d}\boldsymbol{x}_t = \boldsymbol{x}_t\text{d}t + \sqrt{2t}\text{d}\boldsymbol{w}_t\), where \(t \in [0, t_\text{max}]\), and \(\boldsymbol{w}_t\) denotes the Brownian motion. With a sample from the dataset, \(\boldsymbol{x}_0 \sim p_\text{data}(\boldsymbol{x})\), and a random noise, \( \boldsymbol{z} \sim \mathcal{N}(\boldsymbol{z};\boldsymbol{0},\mathbf{I})\), EDM learns the reverse process via the \textit{denoising objective}, 
\begin{equation}
    \mathbb{E} \left[ \lambda(t) \left\| D_{\theta}(\boldsymbol{x}_t) - \boldsymbol{x}_0 \right\|_2^2 \right],
    \label{eq:edm-obj}
\end{equation}
where \(\boldsymbol{x}_t = \boldsymbol{x}_0+t\boldsymbol{z}\), \(\lambda(t)\) is a weighting function w.r.t. \(t\), and \(D_{\theta}\) is further parametrized by \(D_{\theta}(\boldsymbol{x}_t; t) = c_{\text{skip}}(t) \boldsymbol{x}_t + c_{\text{out}}(t) F_{\theta}(c_{\text{in}}(t) \boldsymbol{x}_t; c_{\text{noise}}(t))\), allowing the network to predict residual information from the signal and noise mixture. Besides, the \textit{preconditioning terms}, \(c_{\text{skip}}, c_{\text{out}}, c_{\text{in}}\), ensure unit variance of both the model's target and input. For readers interested in the derivations, please refer to Appendix B.6 from the EDM paper.

\noindent{\bf Discrete-time CM.}
\label{p:discreteCM}We focus our discussion on the family of CMs that model the same ODE as EDM. CMs aim to inference the solution of the PF-ODE at \(t=0\) directly or with few NFEs. To achieve this, CMs are optimized against the \textit{consistency objective},
\begin{equation}
    \mathbb{E}[\lambda(t_{n+1})d(D_{\theta}(\boldsymbol{x}_{t_{n+1}},t_{n+1}),D_{{\theta}^{-}}( \boldsymbol{x}_{t_{n}},t_{n}))],
    \label{eq:cm-obj}
\end{equation}
where \(\{ t_n \}_{n=0}^{N}\) is the discrete timestep sequence defined throughout the course of training, and \(d(\boldsymbol{\cdot},\boldsymbol{\cdot})\) is the distance function. The \textit{noisier term} \(\boldsymbol{x}_{t_{n+1}}\) is given by \(\boldsymbol{x}_0 + t_{n+1} \boldsymbol{z}\), while the \textit{cleaner term} \(\boldsymbol{x}_{t_n}\) can be obtained from the instantaneous change at \(\boldsymbol{x}_{t_{n+1}}\) using a pretrained EDM or by \(\boldsymbol{x}_{t_{n+1}} + (t_n - t_{n+1}) \boldsymbol{z}\), sharing the same noise direction \(\boldsymbol{z}\). The former approach, known as Consistency Distillation (CD), \textit{distills knowledge} from a pretrained EDM, while the latter data-driven method is called Consistency Training (CT). At training, gradient updates are applied only to the weights predicting the noisier term, while the weights for the cleaner term, denoted \(\theta^-\), remains frozen. In CMs, the preconditioning terms look nearly identical to the ones used in EDM to ensure the \textit{boundary condition} of the consistency objective: at \(t=t_\text{min}\), \(D_{\theta}(\boldsymbol{x}) = \boldsymbol{x}\), with \(c_{\text{skip}}(t_\text{min}) = 1\), and \(c_{\text{out}}(t_\text{min}) = 0\), where \(t_\text{min} \in [0,t_\text{max}]\) is the lowest noise at training for stability \cite{song2023consistency}.

\noindent{\bf Continuous-time CM.} We focus on continuous-time CMs introduced in ECT \cite{ect}. ECT is a consistency training/tuning algorithm replacing the discrete timesteps \(\{ t_n \}_{n=0}^{N}\) with a continuous-time schedule. ECT samples noise scales following a lognormal distribution, \(t \sim p(t) = \textit{LogNormal}(P_\textit{mean}, P_\textit{std}), r \sim p(r|t)\) where \(t > r\), optimizing Eq. \ref{eq:cm-obj} between noisy samples \(x_t\) and \(x_r\) along the same noise direction. Throughout training, ECT reduces \(\Delta t:=t-r\) to \(\text{d}t\) via an exponentially decreasing schedule with a base of \(\frac{1}{2}\), e.g., \(\Delta t (t)=\frac{t}{2^{\left\lfloor k/d \right\rfloor}}n(t)\), where \(k, d\) is the training and doubling iterations, and \(n(t)\) is a sigmoid adjusting function. Our iGCT is trained under the same continuous-time schedule used by ECT. To establish DM independence, iGCT's training curriculum begins with diffusion training by setting \(r=t_\text{min}\).

\vspace{-0.15cm}
\subsection{Classifier-free Guidance}
\vspace{-0.15cm}

CFG is a technique in DMs that improves image quality by jointly optimizing an unconditional and conditional denoising objective \cite{ho2022classifier}. At inference, guidance is achieved by extrapolating conditional and unconditional passes using factor \(w\).
\begin{equation}
\frac{\text{d}\boldsymbol{x}}{\text{d}t} = \frac{-\left(w D_{\theta}(\boldsymbol{x}|c;t) + (1-w)D_{\theta}(\boldsymbol{x}|\emptyset;t) - \boldsymbol{x}\right)}{t},
\end{equation}
when \(w>1\), the update direction is encouraged by the conditional term, \(c\), and discouraged from the unconditional term, \(\emptyset\). In practice, high guidance enables high-fidelity image generation \cite{rombach2022high,podell2023sdxlimprovinglatentdiffusion,dieleman2022guidance,ho2022imagenvideohighdefinition,Meng_2023_CVPR} and precise image editing \cite{mokady2023null,miyake2023negative,han2024proxedit,garibi2024renoise,huberman2024edit,starodubcev2024invertible}. High CFG guidance causes deviation from the original PF-ODE, leading to mode drop and overly saturated images \cite{saharia2022photorealistictexttoimagediffusionmodels,kynkäänniemi2024applyingguidancelimitedinterval,bradley2024classifierfreeguidancepredictorcorrector}. Further comparisons of guidance are provided in Sec. \ref{sec:method-gct} and Fig. \ref{fig:1d_cfg_igct}.

\noindent{\bf Guidance in Latent Consistency Models.}
Latent CMs \cite{luo2023latent} are distilled from latent DMs that synthesizes high-resolution images requiring 2-4 NFEs. At distillation, the cleaner term \(x_{t_n}\) relies on an ODE solver, \(\Psi\), a teacher DM, \(D_{\phi}\), to compute the incorporated CFG knowledge into the consistency objective (Eq. \ref{eq:cm-obj}), e.g.,
\vspace{-0.6em}
\begin{equation}
    \begin{aligned}
    \boldsymbol{x}_{t_n} &= \boldsymbol{x}_{t_{n+1}} + (t_n - t_{n+1}) \cdot [w\Psi(\boldsymbol{x}_{t_{n+1}}, t_{n+1}, c; D_{\phi}) \\
    & + (1-w) \Psi(\boldsymbol{x}_{t_{n+1}}, t_{n+1}, \emptyset; D_{\phi}) ],
    \end{aligned}
\label{eq:lcm}
\end{equation}
where the guidance range \(w \in [w_\text{min},w_\text{max}]\) is chosen as a hyper-parameter during distillation. 
\vspace{-0.15cm}
\subsection{Inversion-based Image Editing}
\vspace{-0.15cm}
Inversion-based editing aims to modify subjects in real images by aligning the forward process of diffusion models with the DDIM inversion trajectory~\cite{song2022denoisingdiffusionimplicitmodels}. Previous works achieve this by tuning learnable representations during the editing process, which often requires time-consuming per-image optimization~\cite{dong2023prompt,li2023stylediffusion,mokady2023null}. While BCM \cite{li2024bidirectional}, a CM that demonstrates capabilities in both image generation and inversion, iGCT distinguishes itself by decoupling the inversion module from the generation. This separation allows for the flexibility for guidance generation with iGCT's denoiser. Despite these advancements, achieving inversion-based editing in one-step guided CMs remains an unexplored challenge. As a pioneering work, iGCT addresses this challenge by focusing on category-based editing, offering a novel approach to the problem. Further analysis and results are provided in Section \ref{sec:image-editing}.

%% file: sec/3_method.tex
\begin{figure}[t!]
    \centering
    \includegraphics[width=0.475\textwidth]{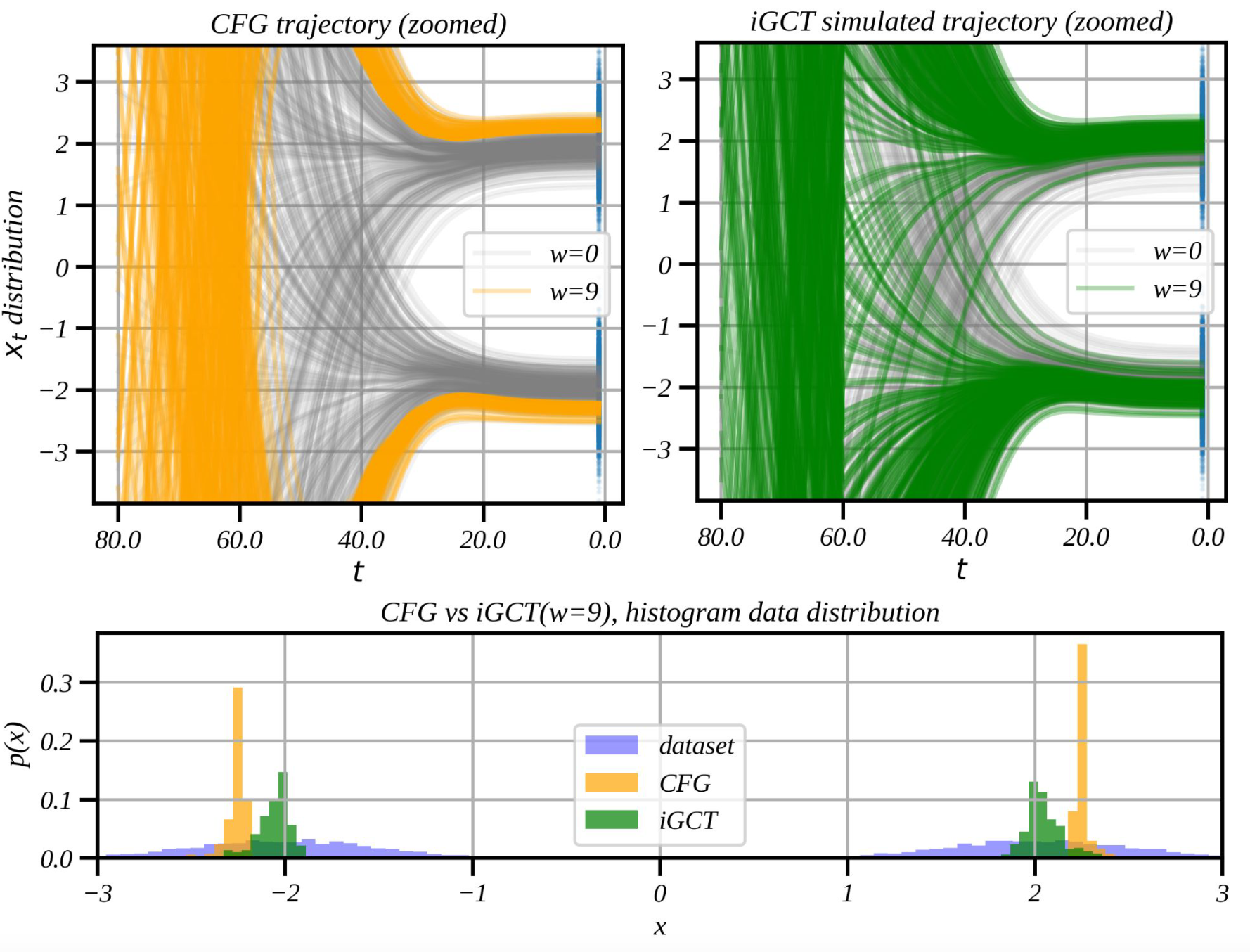} 
    \vspace{-1em}
    \caption{Dynamics of guidance, CFG vs iGCT, 1D toy example with 2 modes at \(x=2\) and \(-2\). Under high guidance, CFG overshoots (in \textcolor{orange}{orange}), \textbf{causing the modes to fall off and producing high contrast values}. Our iGCT (in \textcolor{green}{green}) is able to \textbf{preserve the modes} same as the data distribution (in \textcolor{blue}{blue}). }
    \vspace{-1.5em}
    \label{fig:1d_cfg_igct}  
\vspace{-0.5em}
\end{figure}

\vspace{-0.15cm}
\section{Invertible Guided Consistency Training}
\vspace{-0.15cm}

\begin{figure}[t!]  
    \centering
    \includegraphics[width=0.4\textwidth]{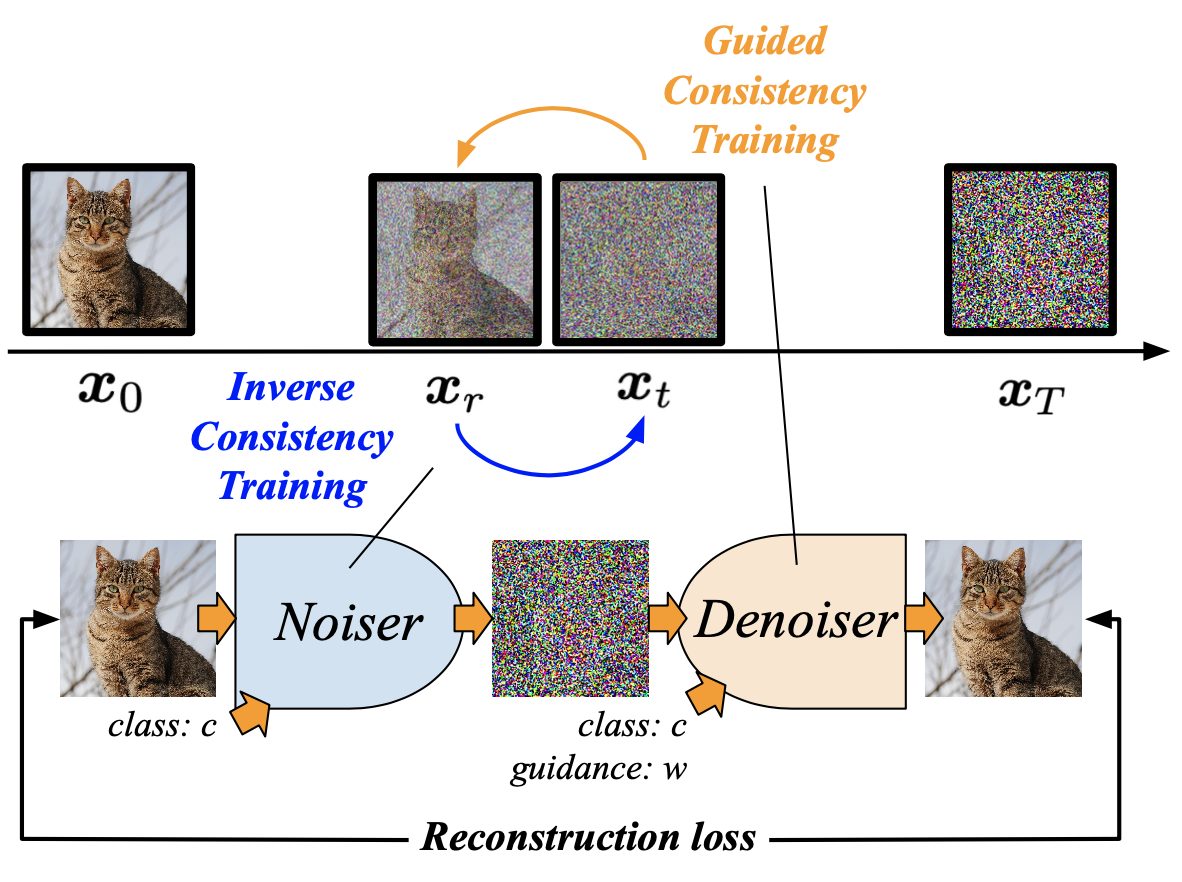} 
    \vspace{-1em}
    \caption{\textbf{Overview of iGCT.} As oppose to the denoiser, the noiser learns to map image to noise by swapping the model's input at training, i.e., the noisier sample \(\boldsymbol{x}_t\) is \(\boldsymbol{x}_r\)'s \textit{target} (See Eq. \ref{eq:inv-loss} for details). The predicted noise latent and denoised image distribution is aligned using the reconstruction loss. }
    \vspace{-1.0em}
    \label{fig:method_overview}  
\end{figure}

\label{sec:method}
This section begins by introducing the approach for data-driven guidance learning with CT, followed by a discussion on the invertible aspect. All together, we present Invertible Guided Consistency Training (iGCT), a one-stage training framework that supports few step, guided image generation and image editing (Fig. \ref{fig:method_overview}). iGCT requires less gpu training hours compared to the total time required for the two-stage framework, i.e., optimizing CFG EDM + guided CD (Table \ref{table:compute_resources}). The resulting iGCT achieves higher precision \cite{kynkäänniemi2019improvedprecisionrecallmetric} compared to a CFG trained DM and a CM distilled from it \cite{song2023consistency,luo2023latent}, allowing better FID under high guidance (Fig. \ref{fig:results_fid_prec_rec}).

\vspace{-0.3cm}
\subsection{The "Curse of Uncondition" and Guided CT}
\label{sec:method-gct}
\vspace{-0.15cm}

For distillation-based CM, guidance is achieved by emphasizing the conditional while negating the unconditional score predicted by a teacher DM. Associated with the input condition \(c\), the same random noise \(\boldsymbol{z}\) that is used to produce the noisy sample \(\boldsymbol{x}_t\) guides toward its original clean image \(\boldsymbol{x}_0\) in conditional CT. Following the notation of continuous-time CM in Sec. \ref{sec:edm_cm}, as \(\Delta t \rightarrow \text{d}t\) with progressively finer discretization during training, the expected instantaneous change from random noise converges to the underlying true conditional noise, denoted as \(\boldsymbol{\epsilon}_{c}\).

\begin{figure}[t!]  
    \centering
    \includegraphics[width=0.475\textwidth]{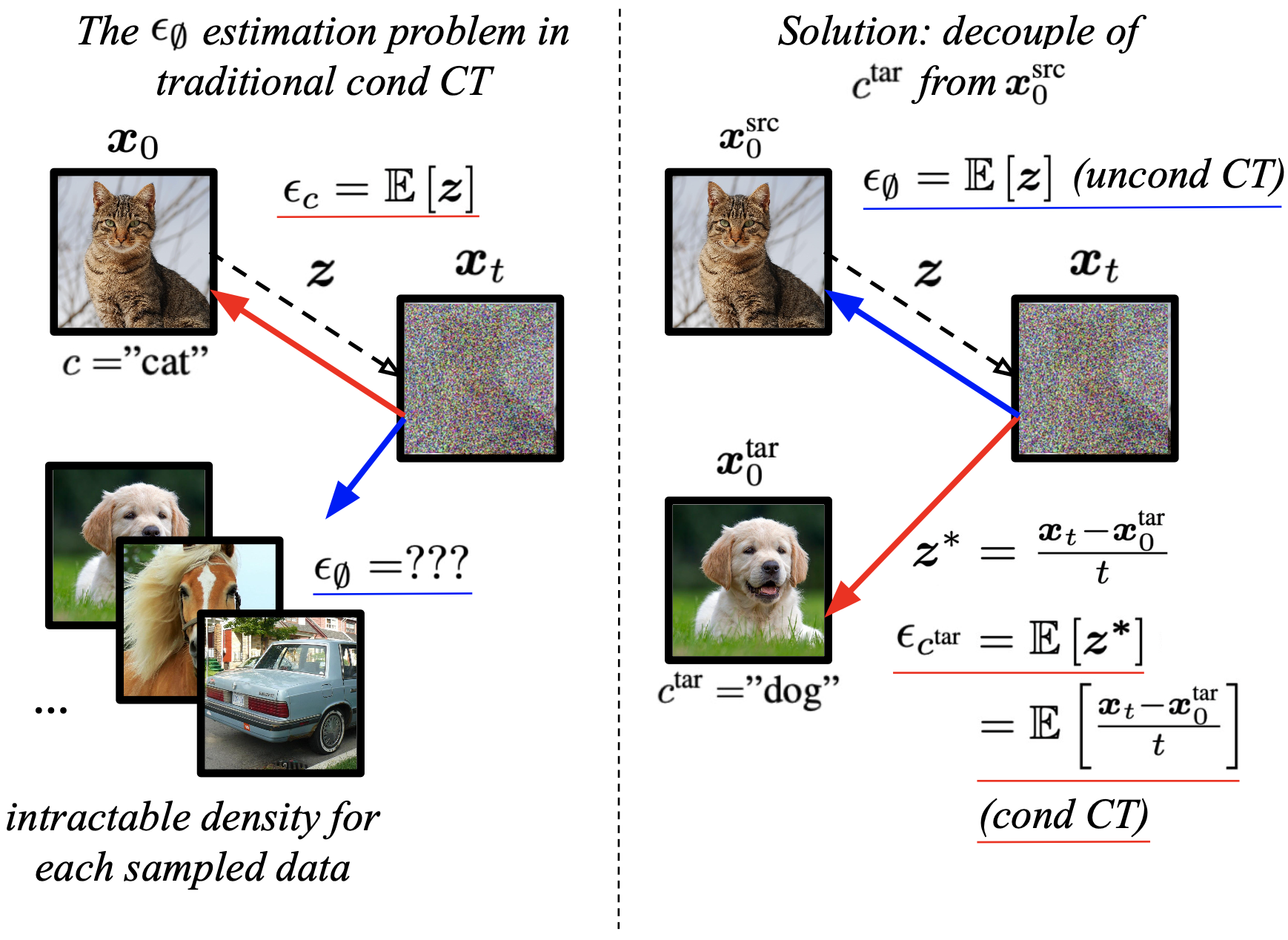} 
    \vspace{-1.2em}
    \caption{In traditional conditional CT, estimating the unconditional noise \(\epsilon_{\emptyset}\) from a noisy sample \(\boldsymbol{x}_t\) requires averaging over all potential clean images (left). By \textbf{decoupling the target class \(c^\text{tar}\) from the source image \(\boldsymbol{x}_0^{\text{src}}\)}, we can estimate \(\epsilon_{\emptyset}\) with \(z\) as how unconditional CT predicts noise, and \(\epsilon_{\text{c}}^{\text{tar}}\) with \(z^*\) like how conditional CT predicts noise (right).}
    \vspace{-1.2em}
    \label{fig:method_gct}  
\end{figure}

However, finding the true unconditional noise is challenging without a DM model. Since guided CMs are defined as \(D_\theta(\boldsymbol{x}_t, t, c, w)\), the sampled noise \(\boldsymbol{z}\) that generated \(\boldsymbol{x}_t\) inherently depends on \(c\), making it unsuitable as an unconditional noise estimate. In other words, \(\boldsymbol{z}\) associates with the class \(c\) inevitably. Prior works address this by training the CM on a denoising objective with the condition masked \cite{hu2024efficienttextdrivenmotiongeneration}, i.e., \(\emptyset\), , allowing it to function as a DM when unconditioned, and switch back to CM when conditioned. To further highlight this challenge, a tempting approach is to approximate the unconditional noise by averaging among random data; however, this requires the conditional density of source samples given \(\boldsymbol{x}_t\), which is effectively what a DM model is trained to model in the absence of class labels.

To address these challenges, we decouple the condition from the source image that generated \(\boldsymbol{x}_t\). We denote \(c^\text{tar}\) as the target condition and \(\boldsymbol{x}_0^{\text{src}}\) as the source image before diffusion, i.e., \(\boldsymbol{x}_t = \boldsymbol{x}_0^{\text{src}} + t \boldsymbol{z}\). As \(\boldsymbol{x}_t\) originates from data independent of \(c^\text{tar}\), we can treat the noise \(\boldsymbol{z}\) as an estimate of the true unconditional noise. The direction from \(\boldsymbol{x}_t\) to a random sample \(\boldsymbol{x}_0^{\text{tar}}\) associated with \(c^\text{tar}\) serves as an estimate of the true conditional noise. Thus, we define the guided direction as the extrapolation between \(\boldsymbol{z}\) and \(\boldsymbol{z}^* := \frac{\boldsymbol{x}_t-\boldsymbol{x}_0^{\text{tar}}}{t}\), yielding \(\boldsymbol{x}_{r} = \boldsymbol{x}_t + (r-t) [w\boldsymbol{z}^*+(1-w)\boldsymbol{z}]\) as the input to the target for the consistency objective (Fig. \ref{fig:method_gct}).

To effectively span the PF-ODE trajectory with sampled data \(\boldsymbol{x}_t\) and \(\boldsymbol{x}_r\) as a coherent \textit{chain}, our loss function incorporates both guided consistency training and the original CT objective (Eq. \ref{eq:cm-obj}) under a continuous-time schedule. Specifically, for high noise levels \(t\), the target \(\boldsymbol{x}_r\) is more likely generated through the above decoupled approach, serving as the target for \(D_\theta(\boldsymbol{x}_t, t, c, w)\) to learn guidance. For low noise samples, \(\boldsymbol{x}_r\) is generated solely via the shared noise \(z\) assuming that \(\epsilon_\emptyset\) and \(\epsilon_{\text{c}}\) are indistinguishable when the sample closely resembles the clean image. We compare the guidance logic of CFG and iGCT using a 1D diffusion toy example with target modes at \(x=2\) and \(-2\) (Fig. \ref{fig:1d_cfg_igct}). CFG shows an overshooting trajectory and generates extreme values away from the modes. In contrast, iGCT produces modes that align with the data distribution. 
The guided CT loss is formulated as follows:
\begin{equation}
     \mathcal{L}_\text{gct} = \lambda(t) d(D_{\theta}(\boldsymbol{x}_t,t,c,w),D_{{\theta}^-}(\boldsymbol{x}_r,r,c,w)). 
    \label{eq:gct-loss}
\end{equation},
where \(\lambda(t)=1/(t-r)\) is the weighting function defined by the reciprocal of the step \(\Delta t\) following ECT \cite{ect}. 

\vspace{-0.15cm}
\subsection{Inverse Consistency Training}
\vspace{-0.15cm}

\label{sec:method-ict}
Besides few-step guided image generation with CT, we introduce inverse CT, which leverages the consistency objective in reverse to map the target distribution back to Gaussian. This deterministic inverse mapping facilitates precise image editing, traditionally performed via DDIM requiring numerous NFEs. By training on the consistency objective, our model directly infers the PF-ODE endpoint efficiently. We define the model that maps images to noise as the \textit{noiser}, denoted \(N_\varphi\).

Following ECT \cite{ect}, we sample \(t \sim p(t) = \textit{LogNormal}(P_\textit{mean}, P_\textit{std})\), and set \(r = t - \Delta t\) by progressively reducing \(\Delta t\) over training. This approach emphasizes importance sampling of pairs \(\boldsymbol{x}_t\) and \(\boldsymbol{x}_r\) in the lower region, where learning the PF-ODE path is more complex and challenging compared to higher noise levels \cite{karras2022elucidating, song2023improved}. As oppose to ECT, we define the weighting function as \(\lambda(t)' = \Delta t(t) / t_\text{max}\), and treat \(\boldsymbol{x}_t\) as the target for inverse CT, optimizing the objective \(\mathbb{E}[\lambda(t)'d(N_\varphi(\boldsymbol{x}_r,r,c),N_{{\varphi}^{-}}( \boldsymbol{x}_t,t,c))]\). Similar to EDM and CMs, our noiser is further preconditioned. We define \(N_{\varphi}(\boldsymbol{x}_t,t,c) = c_\text{skip}(t)\boldsymbol{x}_t + c_\text{out} F_{\varphi}(c_\text{in}\boldsymbol{x}_t, t, c) \), where \(c_\text{in}(t) = 1/\sqrt{t^2+\sigma_\text{data}^2}\), \(c_\text{out}(t) = t_\text{max} - t\) and \(c_\text{skip}(t) = 1\), serving as the boundary conditions while preserving the unit variance property. The proof of unit variance is provided in Appendix \ref{appendix:unit-variance}. The inverse CT loss is given by:
\begin{equation}
     \mathcal{L}_\text{inv} = \lambda(t)'d(N_{\varphi}(\boldsymbol{x}_r,r,c),N_{{\varphi}^-}(\boldsymbol{x}_t,t,c)). 
    \label{eq:inv-loss}
\end{equation}
Following iCD \cite{starodubcev2024invertible}, we found that adding a reconstruction loss is critical to aligning the noiser's latent output with the denoiser's input noise. The reconstruction loss is defined as:
\begin{equation}
     \mathcal{L}_\text{recon} = d(D_{\theta}(N_{\varphi}(\boldsymbol{x}_0,t_\text{min},c),t_\text{max},c,0), \boldsymbol{x}_0),
    \label{eq:recon-loss}
\end{equation}
Putting them all together, the summarized loss terms for our invertible Guided Consistency Training is, \(\mathcal{L} = \mathcal{L}_\text{gct} + \mathcal{L}_\text{inv} + \lambda_{\text{recon}}\mathcal{L}_\text{recon} \). Please refer to Appendix \ref{appendix:iGCT} and Table \ref{tab:igct_training_configs} for the full training algorithm and hyperparameters.

\begin{figure*}[t] 
    \centering
    \includegraphics[width=0.94\textwidth]{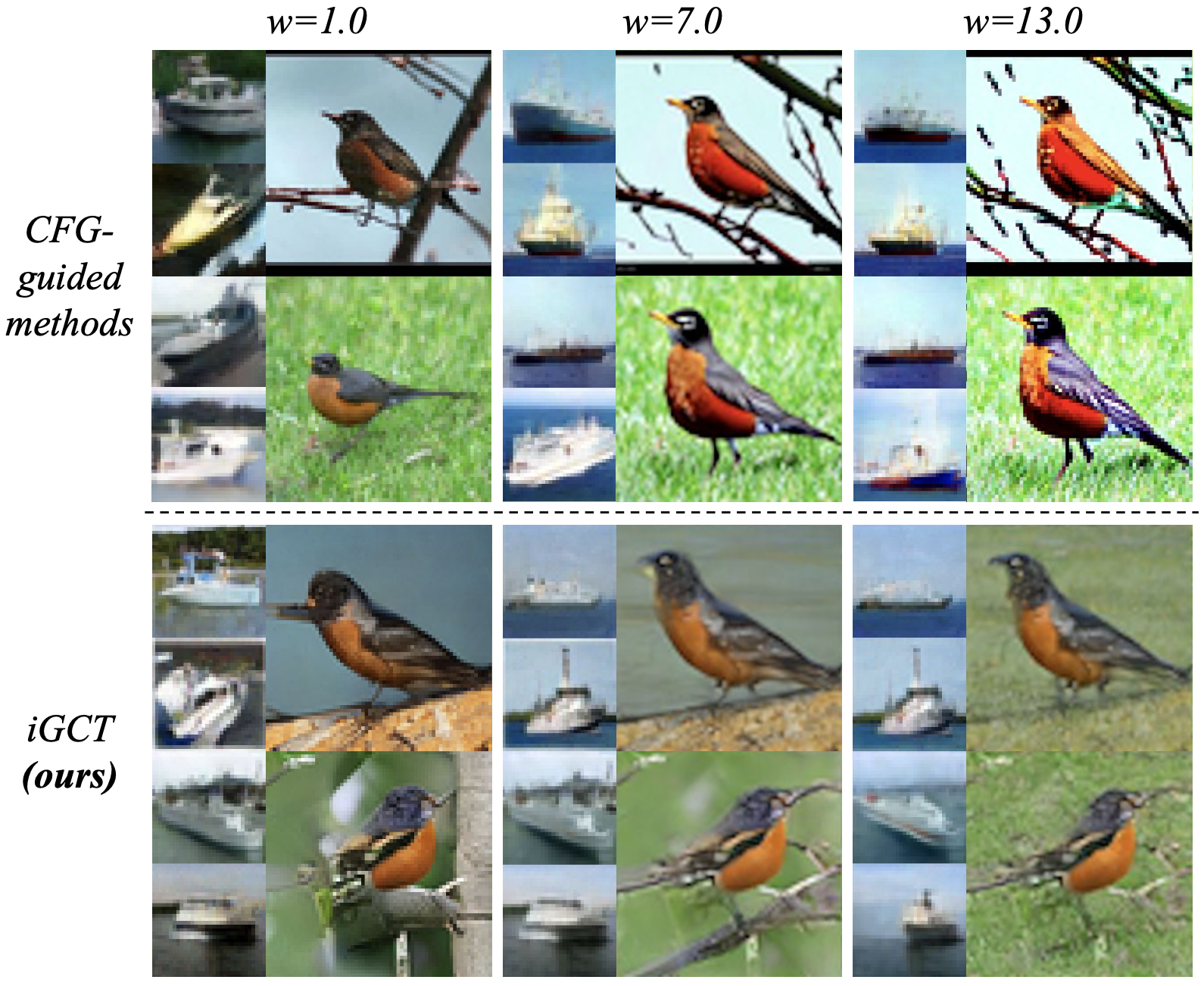} 
    \vspace{-1.3em}
    \caption{Guided generation across levels of guidance scale for CIFAR-10 \textit{"ship"} and ImageNet64 \textit{"robin"}. The top shows images generated via CFG-based methods (\textit{"ship"}: Guided-CD, 1 NFE, \textit{"robin"}: CFG-EDM, 18 NFEs), the below are results generated from our iGCT, 1 NFE. CFG tends to produce high-contrast colors at higher guidance levels, while \textbf{iGCT maintains a consistent overall tone like \(w=0\)} without altering the style.}
    \vspace{-1.5em}
    \label{fig:results_guidance}
\end{figure*}

%% file: sec/4_results.tex
\vspace{-0.2cm}
\section{Experiments: Fast Style-Preserving Guided Generation and Editing}
\label{sec:results} 
\vspace{-0.1cm}

To understand the effect of guidance and inversion of iGCT, we first introduce our baselines, then, present a series of experiments on CIFAR-10 \cite{article} and ImageNet64 \cite{chrabaszcz2017downsampledvariantimagenetalternative} across various guidance scales \(w\) for image generation (Sec. \ref{sec:guidance}) and class-based editing (Sec. \ref{sec:image-editing}). Our analysis mainly focuses on CIFAR-10 compared against CFG and few-step distillation methods as baselines. For training and implementation details on iGCT, please see Appendix \ref{appendix:bs-config}.

\noindent{\bf Baselines.} We compare iGCT with two key baselines: EDM \cite{karras2022elucidating} and CD \cite{song2023consistency}, both widely adopted frameworks for DMs and CMs respectively. The goal is to evaluate iGCT’s performance against multi-step classifier-free guidance (CFG) in DMs, as well as few-step guided CMs. Prior to our work, guidance in consistency models was exclusively achieved through consistency distillation from pretrained DMs, as demonstrated by guided-CD and iCD \cite{luo2023latent,starodubcev2024invertible}. As of writing this paper, iGCT is the first framework to incorporate guidance directly into consistency training, eliminating the need for distillation. Given this, EDM and guided-CD serve as our primary baselines for evaluating guidance performance on CIFAR-10 \cite{article}. Additionally, we conduct image editing experiments using EDM as a baseline by leveraging its invertibility and guidance capabilities. 

We also present results on ImageNet64 \cite{chrabaszcz2017downsampledvariantimagenetalternative}, using EDM as the primary baseline. Due to resource constraints, we exclude guided-CD from this comparison, as distilling a DM model for guided-CD would require approximately twice the computational cost of iGCT from our CIFAR-10 experiments (see Table \ref{table:compute_resources}). This estimate is based on implementing guided-CD following the best configurations outlined in \cite{song2023consistency} and \cite{luo2023latent}. For iGCT, we adopt a smaller ADM architecture \cite{dhariwal2021diffusionmodelsbeatgans}, reducing the width dimensions compared to the default EDM. This adjustment allows us to lower the burden in training iGCT in 190 GPU days on A100 clusters. Additional details about the reimplementation of guided baselines, including CFG-EDM and guided-CD, can be found in Appendix \ref{appendix:bs-config}.

\vspace{-0.3cm}
\subsection{Guidance}
\label{sec:guidance}
\vspace{-0.15cm}

\begin{figure}[t!]  
    \centering
    \begin{subfigure}[b]{0.44\textwidth}
    \includegraphics[width=\textwidth]{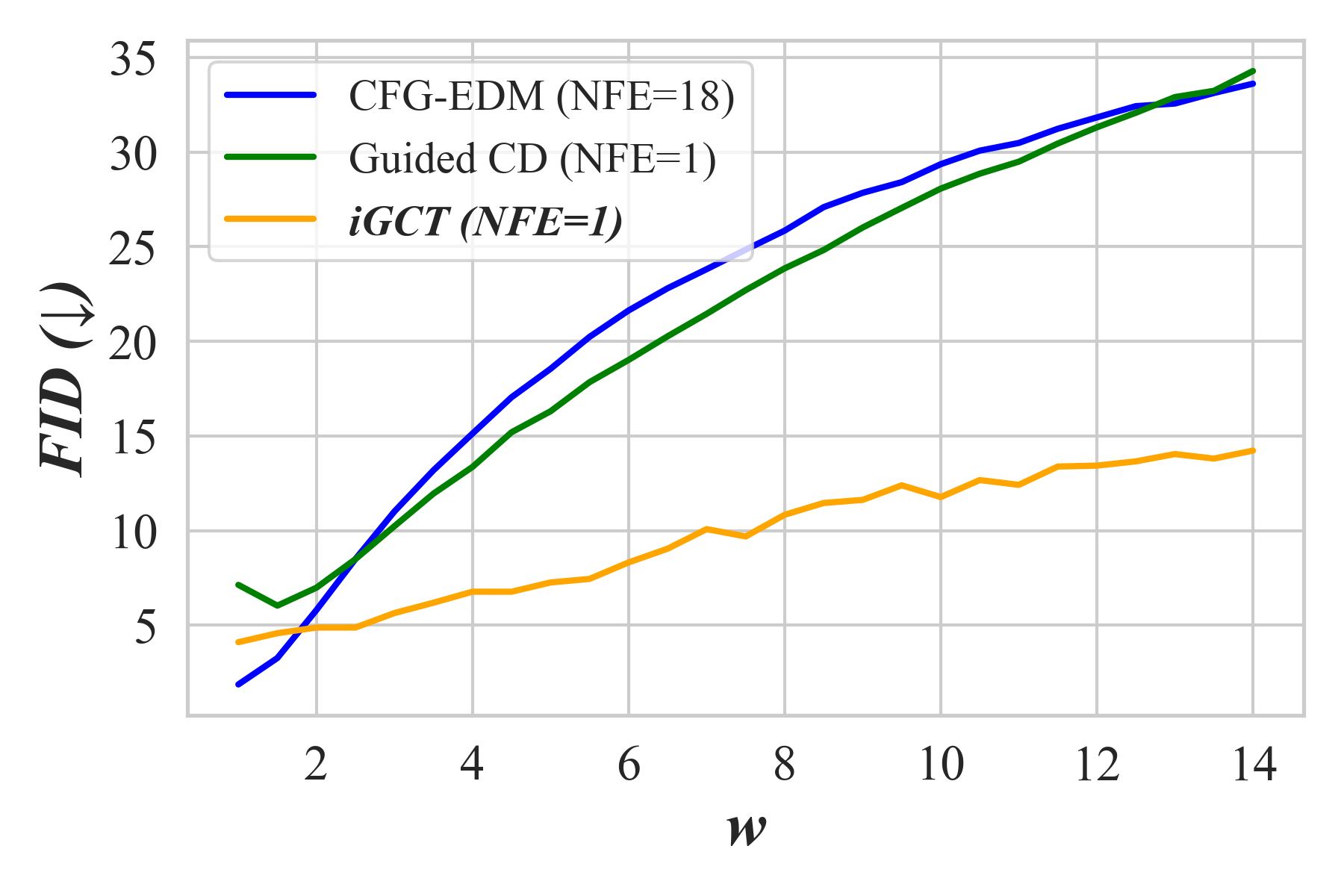} 
        \vspace{-1.4em}
        \caption{FID(50k) on different \(w\) scales.}
    \end{subfigure}
    \hfill
    \begin{subfigure}[b]{0.45\textwidth}
    \includegraphics[width=\textwidth]{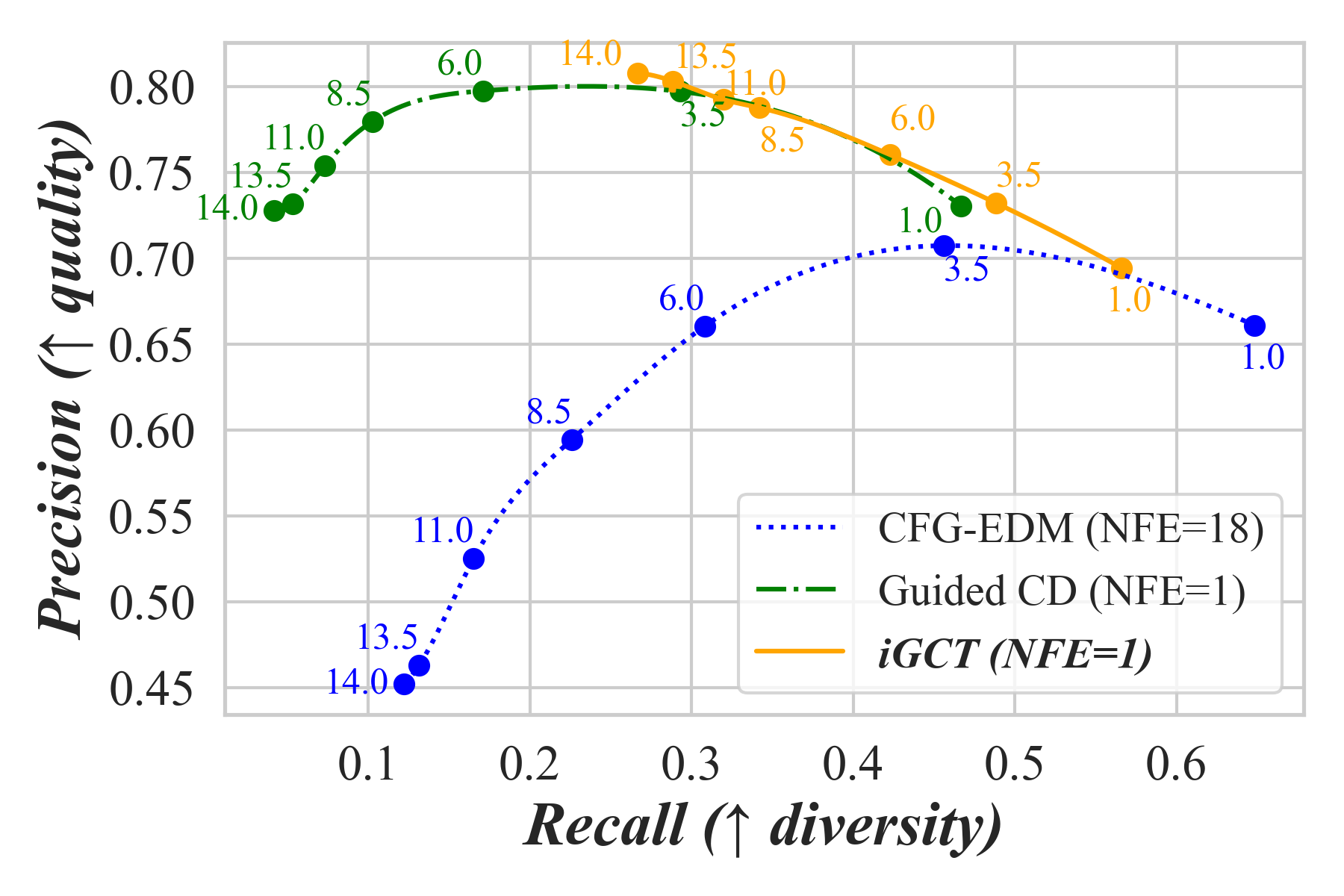} 
        \vspace{-1.4em}
        \caption{Precision/Recall(50k) on different \(w\) scales.}
    \end{subfigure}
    \vspace{-1em}
    \caption{Adjusting the \(w\) scale for iGCT \textbf{consistently enhances precision}, in contrast to CFG, which experiences declines in both quality and diversity beyond a certain threshold. }
    \vspace{-1.5em}
    \label{fig:results_fid_prec_rec}
\end{figure}

\begin{table}[t]
\centering
\caption{Comparison of guided methods using DMs, distillation models, and iGCT, with the \textcolor{blue}{\textbf{best}} score highlighted in blue. iGCT achieves the best precision and FID under high guidance, e.g., w=7, w=13, while also maintaining a strong recall score, showing its ability to stably tradeoff quality and diversity. }
\label{tab:results_metrics}
\vspace{-0.2cm}
\begin{subtable}{0.49\textwidth}
\centering
\caption{CIFAR-10}
\renewcommand{\arraystretch}{0.2}
\resizebox{0.9\textwidth}{!}{
\begin{tabular}{@{}llcc@{}}
    \midrule
    W Scale & Methods & NFE$\downarrow$ & \makecell[l]{FID$\downarrow$ / Precision$\uparrow$ / Recall$\uparrow$} \\
    \midrule
    \multirow{4}{*}{\makecell[l]{w=1}} & \makecell[l]{CFG-EDM \cite{karras2022elucidating}} & 18 & \textcolor{blue}{\textbf{1.9}} / 0.66 / \textcolor{blue}{\textbf{0.64}} \\
    & \makecell[l]{Guided-CD \cite{song2023consistency}} & 1 & 7.1 / \textcolor{blue}{\textbf{0.74}} / 0.47 \\
    & & 2 & 3.9 / 0.71 / 0.53 \\ 
    & \makecell[l]{iGCT (ours)} & 1 & 4.1 / 0.69 / 0.57 \\
    & & 2 & 3.8 / 0.69 / 0.59 \\ 
    \midrule
    \multirow{4}{*}{\makecell[l]{w=7}} & \makecell[l]{CFG-EDM} & 18 & 23.8 / 0.63 / 0.27 \\
    & \makecell[l]{Guided-CD} & 1 & 21.5 / \textcolor{blue}{\textbf{0.79}} / 0.14 \\
    & & 2 & 21.3 / 0.76 / 0.20 \\ 
    & \makecell[l]{iGCT} & 1 & 10.1 / 0.77 / 0.38 \\
    & & 2 & \textcolor{blue}{\textbf{9.2}} / 0.76 / \textcolor{blue}{\textbf{0.42}} \\ 
    \midrule
    \multirow{4}{*}{\makecell[l]{w=13}} & \makecell[l]{CFG-EDM} & 18 & 32.6 / 0.47 / 0.13 \\
    & \makecell[l]{Guided-CD} & 1 & 33.0 / 0.74 / 0.05 \\
    & & 2 & 32.5 / 0.72 / 0.10 \\ 
    & \makecell[l]{iGCT} & 1 & {14.0} / \textcolor{blue}{\textbf{0.80}} / 0.28 \\
    & & 2 & \textcolor{blue}{\textbf{12.6}} / 0.78 / \textcolor{blue}{\textbf{0.35}} \\ 
    \bottomrule
\end{tabular}
}
\end{subtable}%
\vspace{0.1cm}
\begin{subtable}{0.49\textwidth}
\centering
\caption{ImageNet64}
\renewcommand{\arraystretch}{0.2}
\resizebox{0.9\textwidth}{!}{
\begin{tabular}{@{}llcc@{}}
    \midrule
    W Scale & Methods & NFE$\downarrow$ & \makecell[l]{FID$\downarrow$ / Precision$\uparrow$ / Recall$\uparrow$} \\
    \midrule
    \multirow{4}{*}{\makecell[l]{w=1}} & \makecell[l]{CFG-EDM \cite{karras2022elucidating}} & 18 & \textcolor{blue}{\textbf{3.38}} / \textcolor{blue}{\textbf{0.66}} / \textcolor{blue}{\textbf{0.64}} \\
    & \makecell[l]{iGCT} & 1 & 13.16 / 0.46 / 0.40 \\
    & & 2 & 11.67 / 0.43 / 0.47 \\ 
    \midrule
    \multirow{4}{*}{\makecell[l]{w=7}} & \makecell[l]{CFG-EDM} & 18 & 29.19 / \textcolor{blue}{\textbf{0.63}} / 0.27 \\
    & \makecell[l]{iGCT} & 1 & 15.45 / 0.54 / 0.22 \\
    & & 2 & \textcolor{blue}{\textbf{11.18}} / 0.50 / \textcolor{blue}{\textbf{0.29}} \\ 
    \midrule
    \multirow{4}{*}{\makecell[l]{w=13}} & \makecell[l]{CFG-EDM} & 18 & 29.03 / 0.47 / 0.13 \\
    & \makecell[l]{iGCT} & 1 & 20.78 / \textcolor{blue}{\textbf{0.60}} / 0.14 \\
    & & 2 & \textcolor{blue}{\textbf{13.37}} / 0.54 / \textcolor{blue}{\textbf{0.23}} \\ 
    \bottomrule
\end{tabular}
}
\end{subtable}
\vspace{-2.5em}
\end{table}

We evaluate iGCT using FID, precision and recall \cite{kynkäänniemi2019improvedprecisionrecallmetric} on CIFAR-10 and ImageNet64 across various guidance scales \(w\). All three metrics are computed by comparing 50k generated samples with 50k dataset samples. These metrics provide a consistent basis for evaluating a model's performance on quality and diversity. Fig. \ref{fig:results_guidance} compares guidance methods based on CFG and iGCT. When \(w>1\), both CFG-based methods and iGCT exhibit a trade-off between diversity and quality, with increased unification in background tone and color compared to \(w=1\), i.e., unguided conditional generation. Notably, CFG tends to produce high-contrast colors at higher guidance levels, while iGCT maintains a consistent style. Consequently, iGCT achieves lower FID compared to CD and EDM under high guidance. Adjusting the \(w\) scale for iGCT also consistently enhances precision, while CFG shows declines in both quality and diversity beyond a certain threshold. We plot the FID and precision/recall tradeoff on CIFAR-10 in Fig. \ref{fig:results_fid_prec_rec} and provide additional metric results for both CIFAR-10 and ImageNet64 in Table \ref{tab:results_metrics}.

\vspace{-0.15cm}
\subsection{Inversion-based Image Editing}
\vspace{-0.15cm}
\label{sec:image-editing}
To validate iGCT’s effectiveness in image editing, we conduct inversion-based editing experiments on CIFAR-10 and ImageNet64. We compare class-based editing results across various guidance scales \(w\) on the inversed noise latent. We train a CFG-EDM as baseline that infers the noise latent of a source image with DDIM inversion \cite{mokady2023null} without fine-tuning. The edited image is then generated conditionally on a target class. To perform image editing with iGCT, the noiser predicts the noise latent of a source image, then generates the edited image conditioned on the target class. Our baseline EDM requires 18 NFEs for both inversion and generation, whereas iGCT computes the noise latent in a single step, highlighting its potential for real-time image editing.

\begin{figure}[t!]  
    \centering
    \begin{subfigure}[b]{0.475\textwidth}
    \includegraphics[width=\textwidth]{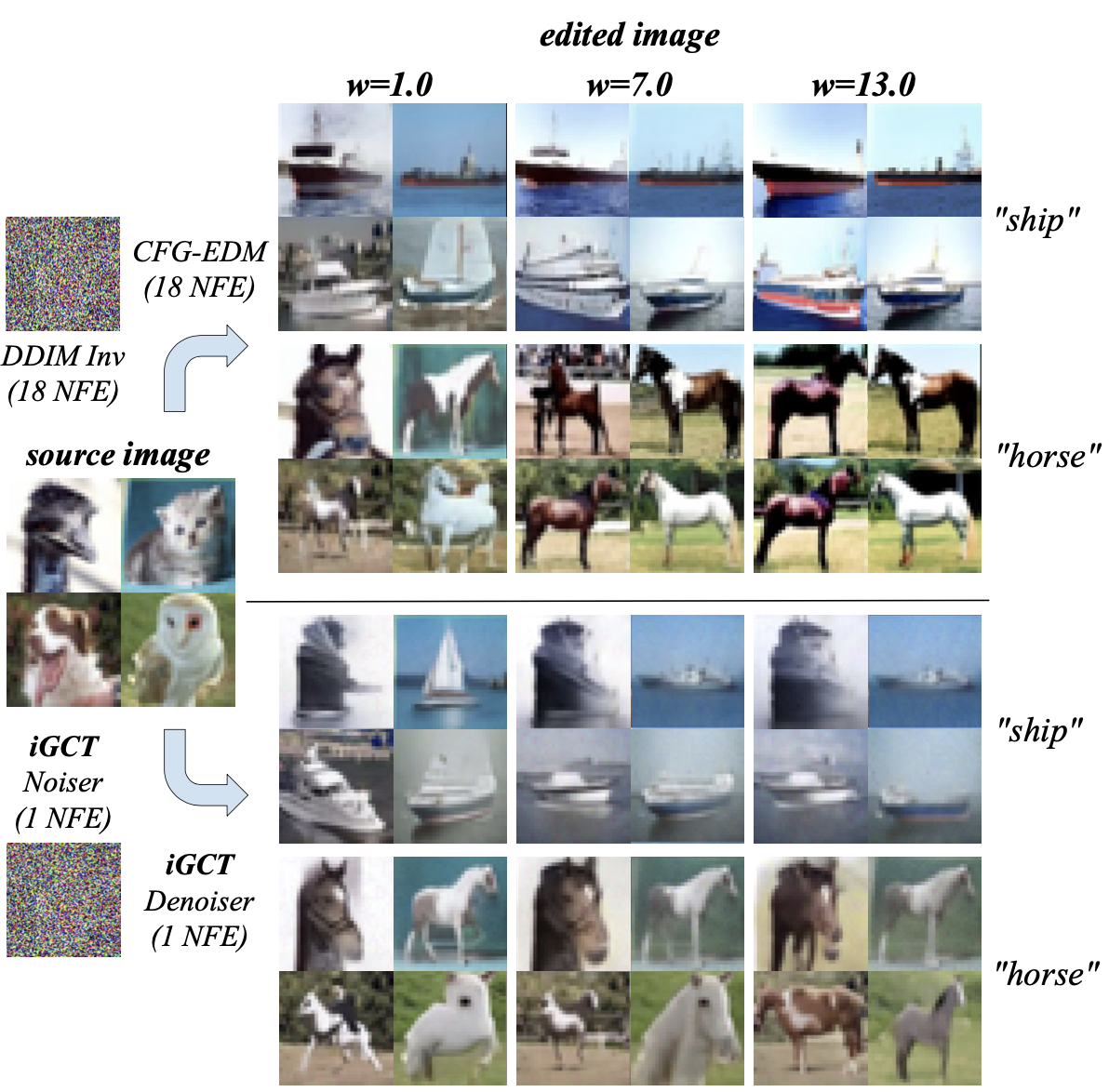} 
        \vspace{-1em}
        \caption{Cross-class image editing on CIFAR-10}
    \end{subfigure}
    \hfill
    \begin{subfigure}[b]{0.45\textwidth}
    \includegraphics[width=\textwidth]{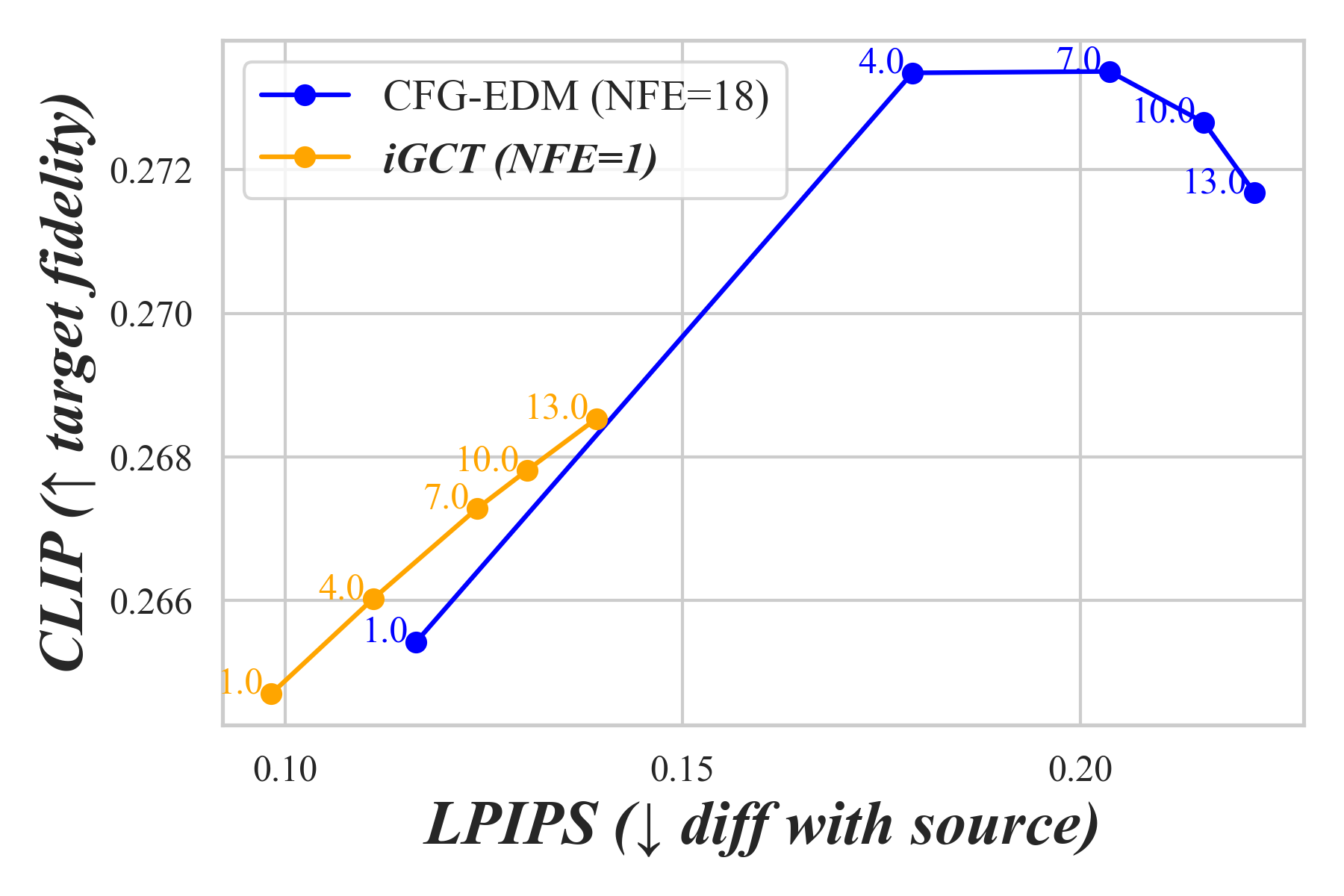} 
        \vspace{-1em}
        \caption{CLIP/LPIPS on different \(w\) scales.}
    \end{subfigure}
    \vspace{-1.1em}
    \caption{iGCT presents strong potential for real-time image editing using CMs. Compared to DM-based methods, iGCT \textbf{aligns source semantics well} and achieves \textbf{rapid edits requiring only a single step.}}
    \label{fig:results_editing}
\vspace{-0.5cm}
\end{figure}

For CIFAR-10, we perform cross-class editing by transforming a source image into each target class. With LPIPS, edits are evaluated by measuring \textit{how much feature is preserved from the source}. With CLIP, we measure \textit{the edit's alignment} with the target prompt \textit{"a photo of a \(\{target\_class\_name\}\)"}. We average LPIPS and CLIP scores across edits for each \(w\) and plot the metrics for both iGCT and the baseline in 2D, illustrating the effects of guidance strength in image editing (Fig. \ref{fig:results_editing}). While EDM achieves a higher CLIP score, it alters the style of the original dataset and deviates from the source image. This shift makes the edited result less relevant to the original content, not to mention its cost in inversion and generation. iGCT presents strong potential for fast inversion-based editing by aligning source semantics well and achieving rapid edits in a single step.

\begin{figure}[t!]  
    \centering
    \begin{subfigure}[b]{0.39\textwidth}
    \includegraphics[width=\textwidth]{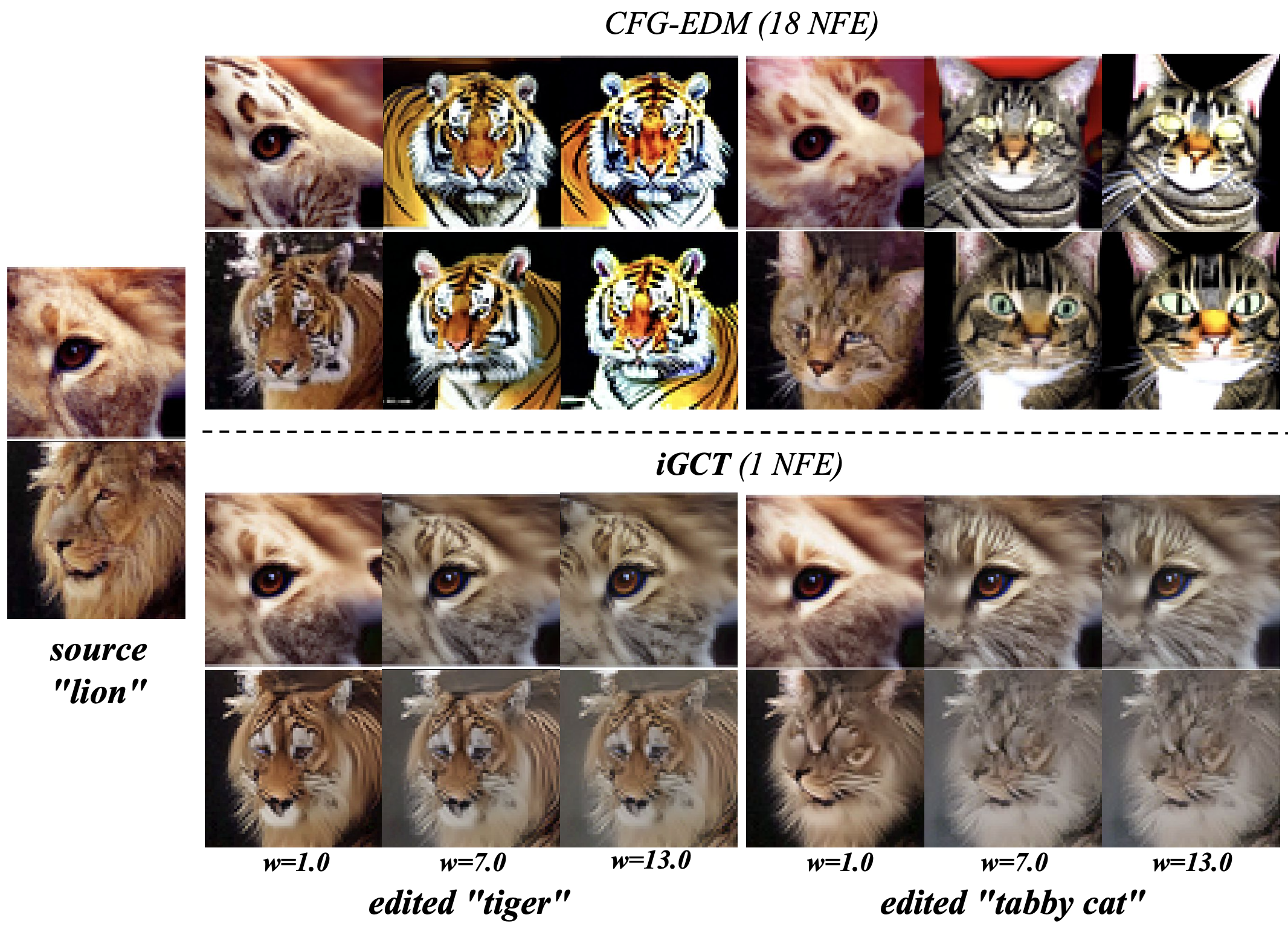} 
        \vspace{-1em}
        \caption{Subgroup: ``cats", CFG-EDM vs iGCT.}
    \end{subfigure}
    
    \begin{subfigure}[b]{0.39\textwidth}
    \includegraphics[width=\textwidth]{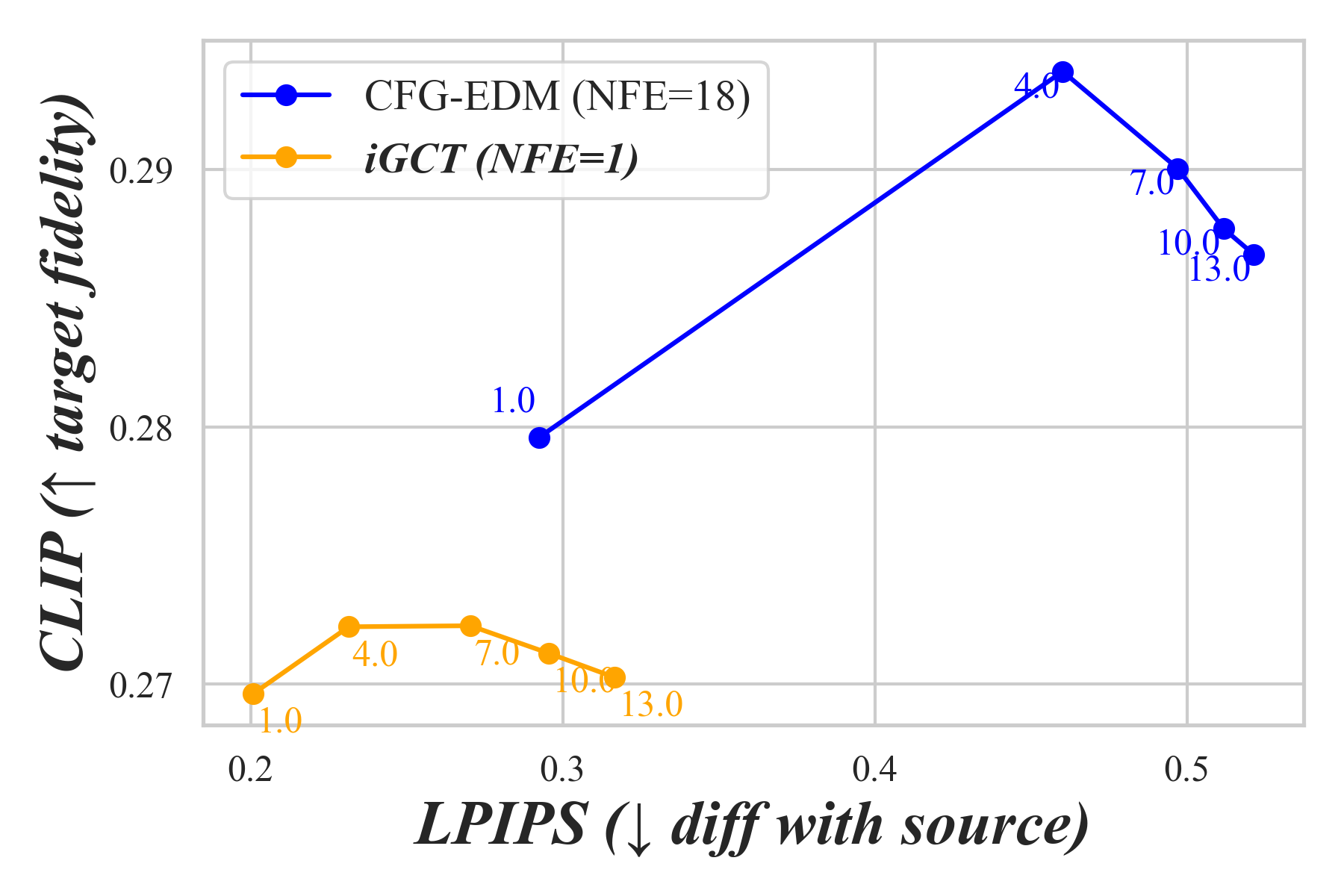} 
        \vspace{-1em}
        \caption{Subgroup: ``cats", CLIP/LPIPS on w scales.}
    \end{subfigure}
    
    \begin{subfigure}[b]{0.39\textwidth}
    \includegraphics[width=\textwidth]{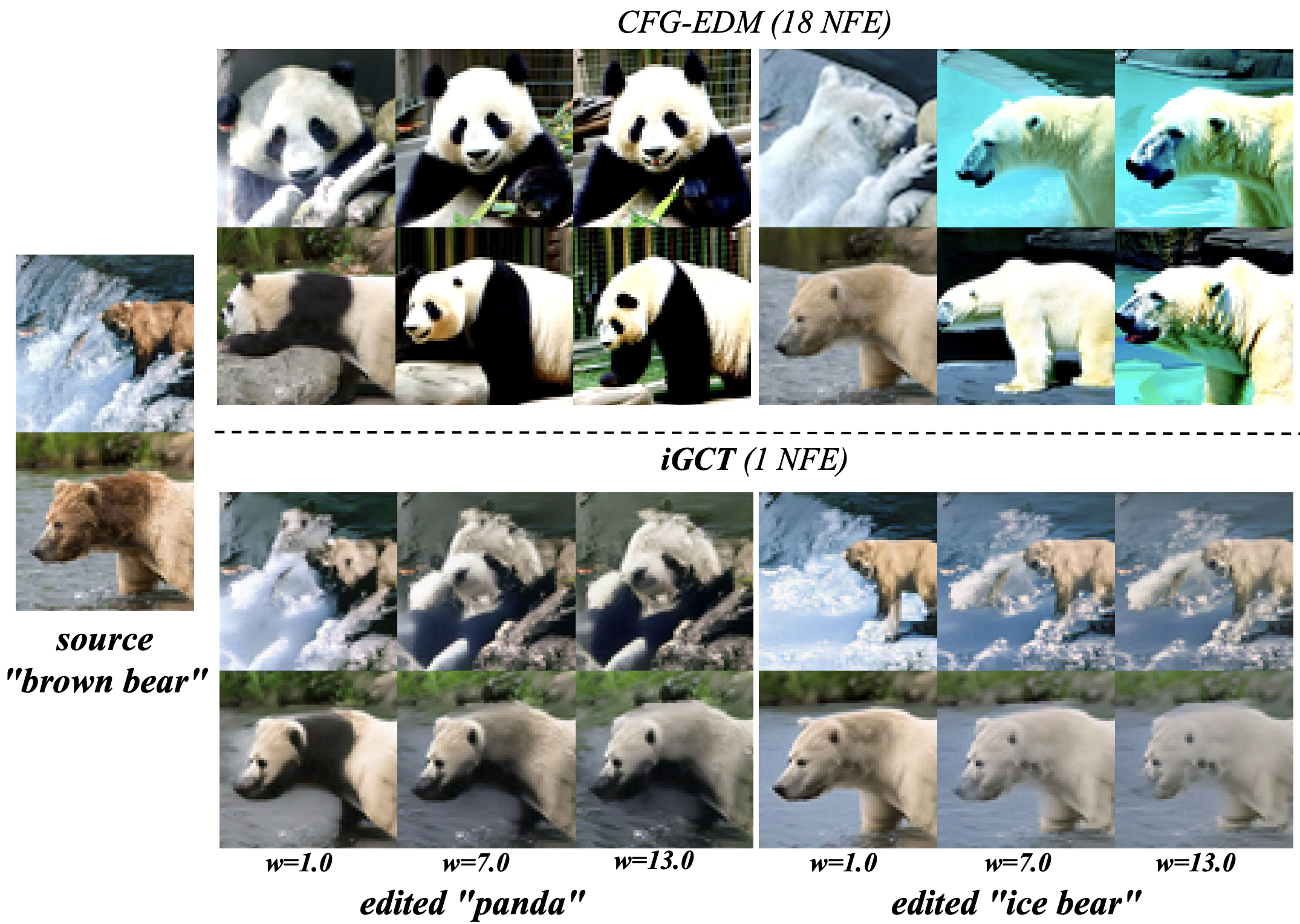} 
        \vspace{-1em}
        \caption{Subgroup:``bears", CFG-EDM vs iGCT.}
    \end{subfigure}
    
    \begin{subfigure}[b]{0.39\textwidth}
    \includegraphics[width=\textwidth]{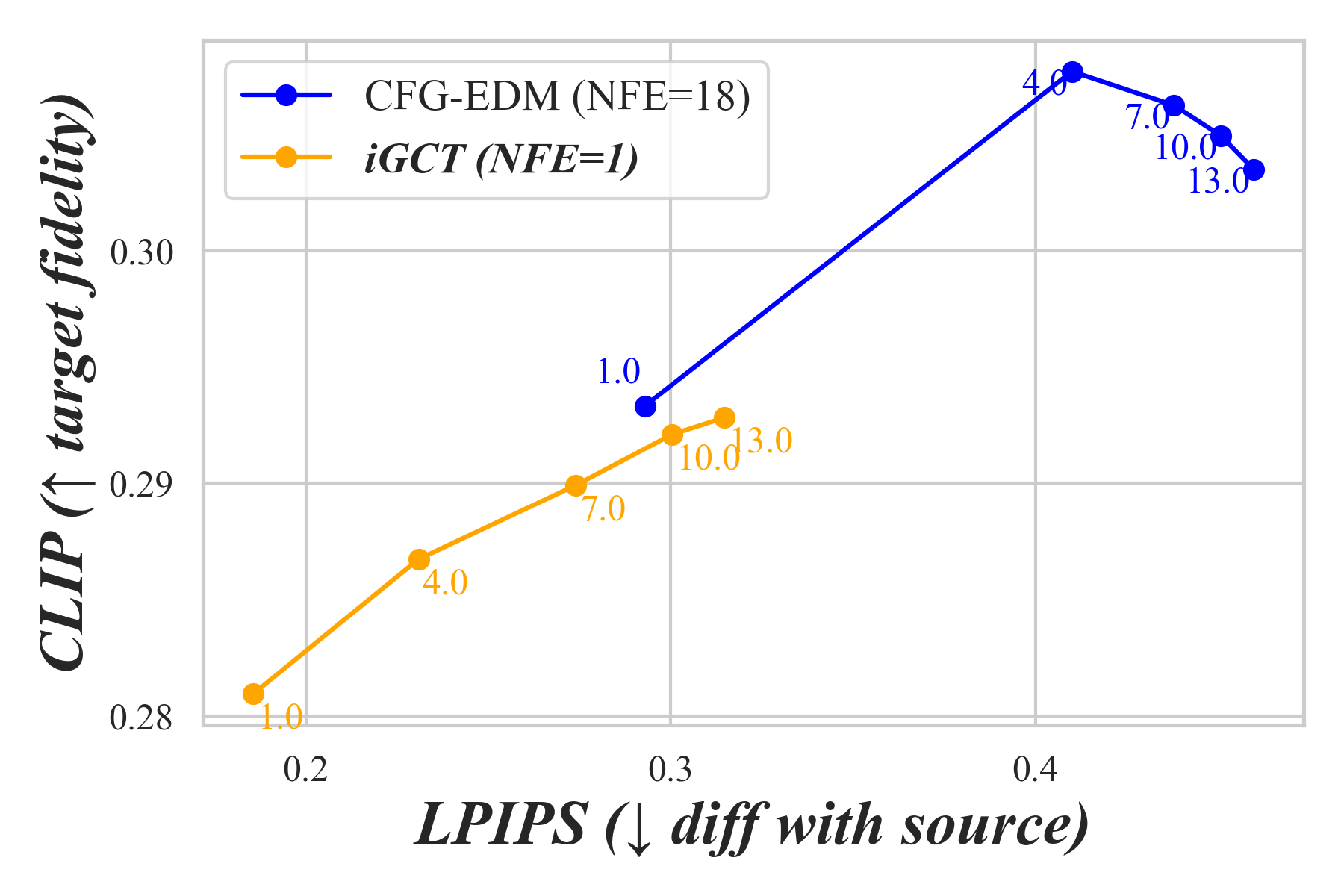} 
        \vspace{-1em}
        \caption{Subgroup: ``bears", CLIP/LPIPS on w scales.}
    \end{subfigure}
    \vspace{-1.1em}
    \caption{Comparison of image editing on ImageNet64 subgroups: "cats" and "bears". iGCT requires only 1 step, and aligns more features from the source compared to EDM. }
    \vspace{-1.3em}
    \label{fig:results_im64_image_editing}
\end{figure}

With 1,000 classes in ImageNet64, we evaluate cross-class editing within 6 predefined subgroups: automobiles, bears, cats, dogs, vegetables, and wild herbivores from the validation set. For image-editing within each subgroup, we chose 5 distinct classes, e.g., black bear is a class in the bears subgroup, and 50 images per class. Results are shown in Fig. \ref{fig:results_im64_image_editing}, where LPIPS and CLIP scores are plotted on different \(w\) for the editing tasks. Similar to findings on CIFAR-10, CFG drastically alters the style and semantics of the source image under high guidance. In contrast, iGCT aligns essential features with both the source and target classes.

%% file: sec/5_conclusion.tex
\vspace{-0.3cm}
\section{Conclusion and Limitation}
\vspace{-0.1cm}

Our proposed iGCT reduces color saturation and artifacts at high guidance levels, offering a data-driven solution for guided image generation in CMs without relying on DMs. We also introduce a novel \textit{noiser} component for efficient image-to-noise mapping, that also enhances the alignment between edited and source images in image editing compared to naive DDIM.

However, iGCT faces several limitations that warrant further exploration. The performance on ImageNet64, for instance, falls short of existing approaches. Additionally, the theory for guided consistency training remains intuitive and informal. Establishing a theoretical mathematical formulation would help clarify why it outperforms alternatives like CFG. Addressing these areas in future work would solidify the method's contributions and open new research directions in this domain. 

\label{sec:conclusion}

%% file: sec/6_appendix.tex
\clearpage
\renewcommand{\thefigure}{A\arabic{figure}}
\renewcommand{\thetable}{A\arabic{table}}
\renewcommand{\theequation}{A\arabic{equation}}
\setcounter{figure}{0}
\setcounter{table}{0}
\setcounter{equation}{0}

Our Appendix is organized as follows. First, we present the pseudocode for the key components of iGCT. We also include the proof for unit variance and boundary conditions in preconditioning iGCT's noiser. Next, we detail the training setups for our CIFAR-10 and ImageNet64 experiments. Additionally, we provide ablation studies on using guided synthesized images as data augmentation in image classification. Finally, we present more uncurated results comparing iGCT and CFG-EDM on inversion, editing and guidance, thoroughly of iGCT.

\vspace{-0.2cm}
\label{appendix:iGCT}
\section{Pseudocode for iGCT}
\vspace{-0.2cm}

iGCT is trained under a continuous-time scheduler similar to the one proposed by ECT \cite{ect}. Our noise sampling function follows a lognormal distribution, \(p(t) = \textit{LogNormal}(P_\textit{mean}, P_\textit{std})\), with \(P_\textit{mean}=-1.1\) and \( P_\textit{std}=2.0\). At training, the sampled noise is clamped at \(t_\text{min} = 0.002\) and \(t_\text{max} = 80.0\). Step function \(\Delta t (t)=\frac{t}{2^{\left\lfloor k/d \right\rfloor}}n(t)\), is used to compute the step size from the sampled noise \(t\), with \(k,d\) being the current training iteration and the number of iterations for halfing \(\Delta t\), and \(n(t) = 1 + 8 \sigma(-t)\) is a sigmoid adjusting function. 

In Guided Consistency Training, the guidance mask function determines whether the sampled noise \( t \) should be supervised for guidance training. With probability \( q(t) \in [0,1] \), the update is directed towards the target sample \( \boldsymbol{x}_0^{\text{tar}} \); otherwise, no guidance is applied. In practice, \( q(t) \) is higher in noisier regions and zero in low-noise regions, 
\begin{equation}
    q(t) = 0.9 \cdot \left( \text{clamp} \left( \frac{t - t_{\text{low}}}{t_{\text{high}} - t_{\text{low}}}, 0, 1 \right) \right)^2,
\end{equation}
where \( t_{\text{low}} = 11.0 \) and \( t_{\text{high}} = 14.3 \). For the range of guidance strength, we set \(w_\text{min} = 1\) and \(w_\text{max} = 15\). Guidance strengths are sampled uniformly at training, with \(w_\text{min} = 1\) means no guidance applied.

\begin{algorithm}
\caption{Guided Consistency Training}
\label{alg:GCT}
\begin{algorithmic}[1]  
\setlength{\baselineskip}{0.9\baselineskip} 
\INPUT Dataset $\mathcal{D}$, weighting function $\lambda(t)$, noise sampling function $p(t)$, noise range $[t_\text{min}, t_\text{max}]$, step function $\Delta t(t)$, guidance mask function $q(t)$, guidance range $[w_\text{min}, w_\text{max}]$, denoiser $D_\theta$
\STATE \rule{0.96\textwidth}{0.45pt} 
\STATE Sample $(\boldsymbol{x}_0^{\text{src}}, c^{\text{src}}), (\boldsymbol{x}_0^{\text{tar}}, c^{\text{tar}}) \sim \mathcal{D}$ 
\STATE Sample noise $\boldsymbol{z} \sim \mathcal{N}(\boldsymbol{0},\mathbf{I})$, time step $t \sim p(t)$, and guidance weight $w \sim \mathcal{U}(w_\text{min}, w_\text{max})$
\STATE Clamp $t \leftarrow \text{clamp}(t,t_\text{min}, t_\text{max})$
\STATE Compute noisy sample: $\boldsymbol{x}_t = \boldsymbol{x}_0^{\text{src}} + t\boldsymbol{z}$
\STATE Sample $\rho \sim \mathcal{U}(0,1)$  
\vspace{0.3em}
\IF{$\rho > q(t)$}
    \STATE Compute step as normal CT: $\boldsymbol{x}_r = \boldsymbol{x}_t - \Delta t(t) \boldsymbol{z}$
    \STATE Set target class: $c \leftarrow c^{\text{src}}$
\ELSE
    \STATE Compute guided noise: $\boldsymbol{z}^* = (\boldsymbol{x}_t - \boldsymbol{x}_0^{\text{tar}}) / t$
    \STATE Compute guided step: $\boldsymbol{x}_r = \boldsymbol{x}_t - \Delta t(t) [w \boldsymbol{z}^* + (1-w)\boldsymbol{z}]$
    \STATE Set target class: $c \leftarrow c^{\text{tar}}$
\ENDIF
\vspace{0.3em} 
\STATE Compute loss: 
\[
\mathcal{L}_\text{gct} = \lambda(t) \, d(D_{\theta}(\boldsymbol{x}_t, t, c, w), D_{{\theta}^-}(\boldsymbol{x}_r, r, c, w))
\]
\STATE Return $\mathcal{L}_\text{gct}$ 
\end{algorithmic}
\end{algorithm}

A \textit{noiser} trained under \textit{Inverse Consistency Training} maps an image to its latent noise in a single step. In contrast, DDIM Inversion requires multiple steps with a diffusion model to accurately produce an image's latent representation. Since the boundary signal is reversed, spreading from \( t_\text{max} \) down to \( t_\text{min} \), we design the importance weighting function \( \lambda'(t) \) to emphasize higher noise regions, defined as:
\begin{equation}
    \lambda'(t) = \frac{\Delta t (t)}{t_\text{max}},
\end{equation}
where the step size \( \Delta t (t) \) is proportional to the sampled noise level \(t\), and \( t_\text{max} \) is a constant that normalizes the scale of the inversion loss. The noise sampling function \( p(t) \) and the step function \( \Delta t (t) \) used in computing both \(\mathcal{L}_\text{gct}\) and \(\mathcal{L}_\text{ict}\) are the same.

\begin{algorithm}
\caption{Inverse Consistency Training}
\label{alg:iCT}
\begin{algorithmic}[1]  
\setlength{\baselineskip}{0.9\baselineskip} 
\INPUT Dataset $\mathcal{D}$, weighting function $\lambda'(t)$, noise sampling function $p(t)$, noise range $[t_\text{min}, t_\text{max}]$, step function $\Delta t(t)$, noiser $N_\varphi$
\STATE \rule{0.96\textwidth}{0.45pt} 
\STATE Sample $\boldsymbol{x}_0, c \sim \mathcal{D}$ 
\STATE Sample noise $\boldsymbol{z} \sim \mathcal{N}(\boldsymbol{0},\mathbf{I})$, time step $t \sim p(t)$
\STATE Clamp $t \leftarrow \text{clamp}(t,t_\text{min}, t_\text{max})$
\STATE Compute noisy sample: $\boldsymbol{x}_t = \boldsymbol{x}_0 + t\boldsymbol{z}$
\STATE Compute cleaner sample: $\boldsymbol{x}_r = \boldsymbol{x}_t - \Delta t(t) \boldsymbol{z}$
\vspace{0.3em} 
\STATE Compute loss: 
\[
\mathcal{L}_\text{ict} = \lambda'(t) \, d(N_{\varphi}(\boldsymbol{x}_r, r, c), D_{{\varphi}^-}(\boldsymbol{x}_t, t, c))
\]
\STATE Return $\mathcal{L}_\text{ict}$ 
\end{algorithmic}
\end{algorithm}

Together, iGCT jointly optimizes the two consistency objectives \(\mathcal{L}_\text{gct}, \mathcal{L}_\text{ict}\), and aligns the noiser and denoiser via a reconstruction loss, \(\mathcal{L}_\text{recon}\). To improve training efficiency, \(\mathcal{L}_\text{recon}\) is computed every \(i_\text{skip}\), reducing the computational cost of back-propagation through both the weights of the \textit{denoiser} \(\theta\) and the \textit{noiser} \(\varphi\). Alg. \ref{alg:iGCT} provides an overview of iGCT. 

\begin{algorithm}
\caption{iGCT}
\label{alg:iGCT}
\begin{algorithmic}[1]  
\setlength{\baselineskip}{0.9\baselineskip} 
\INPUT Dataset $\mathcal{D}$, learning rate $\eta$, weighting functions $\lambda'(t), \lambda(t), \lambda_{\text{recon}}$, noise sampling function $p(t)$, noise range $[t_\text{min}, t_\text{max}]$, step function $\Delta t(t)$, guidance mask function $q(t)$, guidance range $[w_\text{min}, w_\text{max}]$, reconstruction skip iters $i_\text{skip}$, models $N_\varphi, D_\theta$
\STATE \rule{0.9\textwidth}{0.45pt}  
\STATE \textbf{Init:} $\theta, \varphi$, $\text{Iters} = 0$
\REPEAT
\STATE Do guided consistency training 
\[
\mathcal{L}_\text{gct}(\theta;\mathcal{D},\lambda(t),p(t),t_\text{min},t_\text{max},\Delta t(t),q(t),w_\text{min},w_\text{max})
\]
\STATE Do inverse consistency training
\[
\mathcal{L}_\text{ict}(\varphi;\mathcal{D},\lambda'(t),p(t),t_\text{min},t_\text{max},\Delta t(t))
\]
\IF{$(\text{Iters} \ \% \ i_\text{skip}) == 0$}
\STATE Compute reconstruction loss
\[
\mathcal{L}_\text{recon} = d(D_{\theta}(N_{\varphi}(\boldsymbol{x}_0,t_\text{min},c),t_\text{max},c,0), \boldsymbol{x}_0)
\]
\ELSE
\STATE \[
\mathcal{L}_\text{recon} = 0
\]
\ENDIF
\STATE Compute total loss: 
\[
\mathcal{L} = \mathcal{L}_\text{gct} + \mathcal{L}_\text{ict} + \lambda_{\text{recon}}\mathcal{L}_\text{recon}
\]
\STATE $\theta \leftarrow \theta - \eta \nabla_{\theta} \mathcal{L}, \ \varphi \leftarrow \varphi - \eta \nabla_{\varphi} \mathcal{L}$
\STATE $\text{Iters} = \text{Iters} + 1$
\UNTIL{$\Delta t \rightarrow dt$}
\end{algorithmic}
\end{algorithm}

\vspace{-0.3cm}
\section{Preconditioning for Noiser}
\label{appendix:unit-variance}
\vspace{-0.1cm}

We define 
\begin{equation}
    N_{\varphi}(\boldsymbol{x}_t, t, c) = c_\text{skip}(t) \, \boldsymbol{x}_t + c_\text{out}(t) \, F_{\varphi}(c_\text{in}(t) \, \boldsymbol{x}_t, t, c),
\end{equation}
where \( c_\text{in}(t) = \frac{1}{\sqrt{t^2 + \sigma_\text{data}^2}} \), \( c_\text{skip}(t) = 1 \), and \( c_\text{out}(t) = t_\text{max} - t \). This setup naturally serves as a boundary condition. Specifically:

\begin{itemize}
    \item When \( t = 0 \),
    \begin{equation}
        c_\text{out}(0) = t_\text{max} \gg c_\text{skip}(0) = 1,
    \end{equation}
    emphasizing that the model's noise prediction dominates the residual information given a relatively clean sample.

    \item When \( t = t_\text{max} \),
    \begin{equation}
        N_{\varphi}(\boldsymbol{x}_{t_\text{max}}, t_\text{max}, c) = \boldsymbol{x}_{t_\text{max}},
    \end{equation}
    satisfying the condition that \( N_{\varphi} \) outputs \( \boldsymbol{x}_{t_\text{max}} \) at the maximum time step.
\end{itemize}

We show that these preconditions ensure unit variance for the model’s input and target. First, \(\text{Var}_{\boldsymbol{x}_0, z}[\boldsymbol{x}_t] = \sigma_\text{data}^2 + t^2\), so setting \( c_\text{in}(t) = \frac{1}{\sqrt{\sigma_\text{data}^2 + t^2}} \) normalizes the input variance to 1. Second, we require the training target to have unit variance. Given the noise target for \( N_{\varphi} \) is \(\boldsymbol{x}_{t_\text{max}} = \boldsymbol{x}_0 + t_\text{max} z\), by moving of terms, the effective target for \( F_{\varphi} \) can be written as,
\begin{equation}
    \frac{\boldsymbol{x}_{t_\text{max}} - c_\text{skip}(t)\boldsymbol{x}_{t}}{c_\text{out}(t)}
\end{equation}
When \(c_\text{skip}(t) = 1\), \(c_\text{out}(t) = t_\text{max} - t \), we verify that target is unit variance,
\begin{align}
    &\text{Var}_{\boldsymbol{x}_0, \boldsymbol{z}} \left[ \frac{\boldsymbol{x}_{t_\text{max}} - c_\text{skip}(t) \, \boldsymbol{x}_{t}}{c_\text{out}(t)} \right] \\ \notag
    = \ &\text{Var}_{\boldsymbol{x}_0, \boldsymbol{z}} \left[ \frac{\boldsymbol{x}_0 + t_\text{max} \, \boldsymbol{z} - (\boldsymbol{x}_0 + t \, \boldsymbol{z})}{t_\text{max} - t} \right] \notag \\
    = \ &\text{Var}_{\boldsymbol{x}_0, \boldsymbol{z}} \left[ \frac{(t_\text{max} - t) \, \boldsymbol{z}}{t_\text{max} - t} \right] \notag \\
    = \ &\text{Var}_{\boldsymbol{x}_0, \boldsymbol{z}}[\boldsymbol{z}] \notag \\
    = \ &1. \notag
\end{align}

\vspace{-0.3cm}
\section{Baselines \& Training Details}
\label{appendix:bs-config}
\vspace{-0.1cm}

\begin{figure}[t!]  
    \centering
    \begin{subfigure}[b]{0.33\textwidth}
    \includegraphics[width=\textwidth]{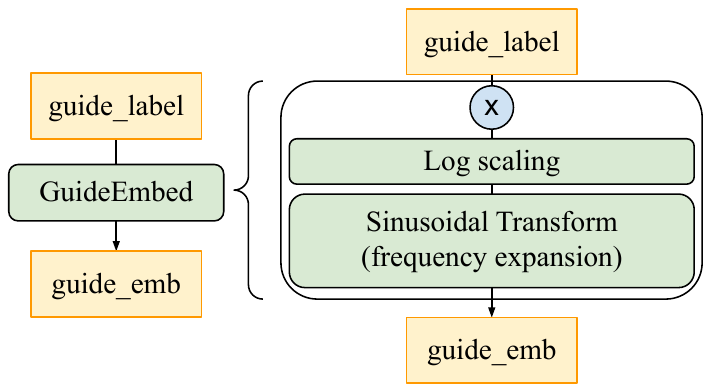} 
        \caption{Guidance embedding.}
    \end{subfigure}
    \hfill
    \begin{subfigure}[b]{0.33\textwidth}
    \includegraphics[width=\textwidth]{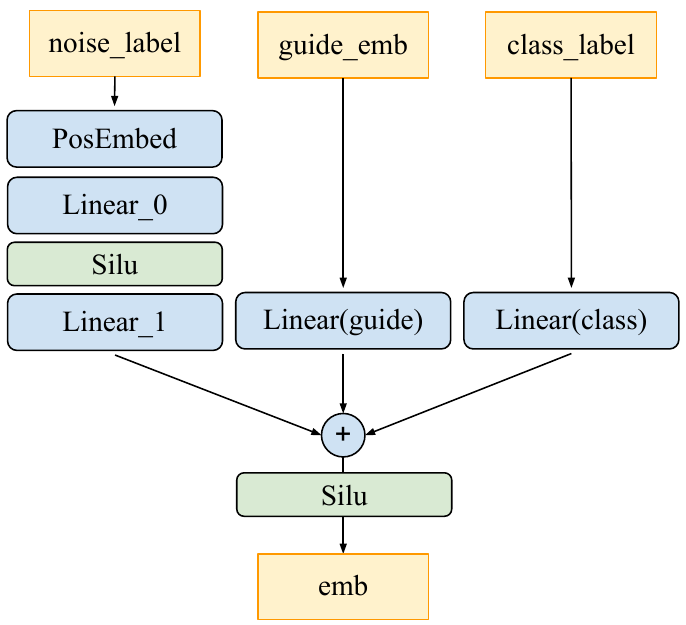} 
        \caption{NCSN++ architecture.}
    \end{subfigure}
    \hfill
    \begin{subfigure}[b]{0.33\textwidth}
    \includegraphics[width=\textwidth]{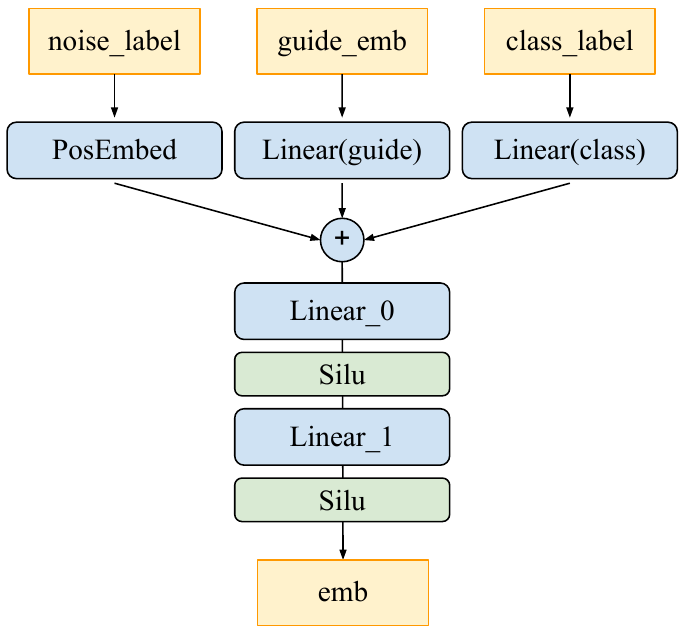} 
        \caption{ADM architecture.}
    \end{subfigure}
    \hfill
    \caption{Design of guidance embedding, and conditioning under different network architectures.}
    \vspace{-1em}
    \label{fig:guidance_conditioning}
\end{figure}

For our diffusion model baseline, we follow \textit{EDM}'s official repository (\href{https://github.com/NVlabs/edm}{https://github.com/NVlabs/edm}) instructions for training and set \textit{label\_dropout} to 0.1 to optimize a CFG (classifier-free guided) DM. We will use this DM as the teacher model for our consistency model baseline via consistency distillation. 

The consistency model baseline \textit{Guided CD} is trained with a discrete-time schedule. We set the discretization steps \( N = 18 \) and use a Heun ODE solver to predict update directions based on the CFG EDM, as in \cite{song2023consistency}. Following \cite{luo2023latent}, we modify the model's architecture and iGCT's denoiser to accept guidance strength \(w\) by adding an extra linear layer. See the detailed architecture design for guidance conditioning of consistency model in Fig. \ref{fig:guidance_conditioning}. A range of guidance scales \(w \in [1,15]\) is uniformly sampled at training. Following \cite{song2023improved}, we replace LPIPS by Pseudo-Huber loss, with \(c=0.03 \) determining the breadth of the smoothing section between L1 and L2. See Table \ref{tab:training_configs} for a summary of the training configurations for our baseline models.

\begin{table}[t!]
\centering
\renewcommand{\arraystretch}{1.3} 
\small 
\caption{Summary of training configurations for baseline models.}
\begin{tabular}{lccc}
\toprule
\multirow{2}{*}{} & \multicolumn{2}{c}{\textbf{CIFAR-10}} & \textbf{ImageNet64}  \\
                  & EDM & Guided-CD & EDM \\
\midrule
\multicolumn{4}{l}{\textbf{\small Arch. config.}} \\
\hline
model arch.        & NCSN++ & NCSN++ & ADM     \\
channels mult.     & 2,2,2  & 2,2,2  & 1,2,3,4 \\
UNet size          & 56.4M  & 56.4M  & 295.9M  \\
\midrule
\multicolumn{4}{l}{\textbf{\small Training config.}} \\
\hline
lr             & 1e-3  & 4e-4  & 2e-4 \\
batch          & 512   & 512   & 4096 \\
dropout        & 0.13  & 0     & 0.1 \\
label dropout  & 0.1   & (n.a.) & 0.1 \\
loss           & L2    & Huber & L2    \\
training iterations & 390k  & 800k  & 800K \\
\bottomrule
\end{tabular}
\label{tab:training_configs}
\end{table}

\begin{table}[t!]
\centering
\renewcommand{\arraystretch}{1.3} 
\small 
\caption{Summary of training configurations for iGCT.}
\begin{tabular}{lcc}
\toprule
\multirow{2}{*}{} & \textbf{CIFAR-10} & \textbf{ImageNet64}  \\
                  & iGCT & iGCT \\
\midrule
\multicolumn{3}{l}{\textbf{\small Arch. config.}} \\
\hline
model arch.        & NCSN++ & ADM \\
channels mult.     & 2,2,2  & 1,2,2,3 \\
UNet size          & 56.4M  & 182.4M \\ 
Total size         & 112.9M & 364.8M \\ 
\midrule
\multicolumn{3}{l}{\textbf{\small Training config.}} \\
\hline
lr              & 1e-4 & 1e-4 \\
batch           & 1024 & 1024 \\
dropout            & 0.2 & 0.3 \\
loss               & Huber   & Huber \\
\(c\)                  & 0.03    &  0.06 \\
\(d\)                  & 40k     &  40k \\
\( P_\textit{mean} \) & -1.1 &  -1.1 \\
\( P_\textit{std} \) &  2.0  &  2.0  \\
\( \lambda_{\text{recon}} \) & 2e-5 & \parbox[t]{3.5cm}{\centering 2e-5, (\(\leq\) 180k)\\ 4e-5, (\(\leq\) 200k)\\ 6e-5, (\(\leq\) 260k) } \\  
\( i_{\text{skip}} \)        & 10 &  10 \\  
training iterations & 360k &  260k \\
\bottomrule
\end{tabular}
\label{tab:igct_training_configs}
\end{table}  

\begin{figure*}[t] 
    \centering
    \includegraphics[width=1.0\textwidth]{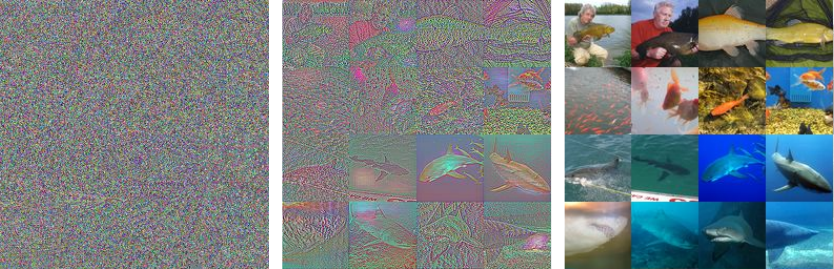} 
    \caption{Inversion collapse observed during training on ImageNet64. The left image shows the input data. The middle image depicts the inversion collapse that occurred at iteration 220k, where leakage of signals in the noise latent can be visualized. The right image shows the inversion results at iteration 220k after appropriately increasing $\lambda_{\text{recon}}$ to 6e-5. The inversion images are generated by scaling the model's outputs by $1/80$, i.e., $ 1/t_\text{max}$, then clipping the values to the range [-3, 3] before denormalizing them to the range [0, 255]. }
    \vspace{-1.5em}
    \label{fig:inversion_collpase}
\end{figure*}

iGCT is trained with a continuous-time scheduler inspired by ECT \cite{ect}. To rigorously assess its independence from diffusion-based models, iGCT is trained from scratch rather than fine-tuned from a pre-trained diffusion model. Consequently, the training curriculum begins with an initial diffusion training stage, followed by consistency training with the step size halved every \(d\) iterations. In practice, we adopt the same noise sampling distribution \(p(t)\), same step function \(\Delta t (t) \), and same distance metric \( d(\cdot, \cdot) \) for both guided consistency training and inverse consistency training. 

For CIFAR-10, iGCT adopts the same UNet architecture as the baseline models. However, the overall model size is doubled, as iGCT comprises two UNets: one for the denoiser and one for the noiser. The Pseudo-Huber loss is employed as the distance metric, with a constant parameter \( c = 0.03 \). Consistency training is organized into nine stages, each comprising 400k iterations with the step size halved from the last stage. We found that training remains stable when the reconstruction weight \( \lambda_{\text{recon}} \) is fixed at \( 2 \times 10^{-5} \) throughout the entire training process.
 
For ImageNet64, iGCT employs a reduced ADM architecture \cite{dhariwal2021diffusionmodelsbeatgans} with smaller channel sizes to address computational constraints. A higher dropout rate and Pseudo-Huber loss with \( c = 0.06 \) is used, following prior works \cite{ect,song2023improved}. During our experiments, we observed that training on ImageNet64 is sensitive to the reconstruction weight. Keeping \(\lambda_{\text{recon}}\) fixed throughout training leads to inversion collapse, with significant signal leaked to the latent noise (see Fig. \ref{fig:inversion_collpase}). We found that increasing \(\lambda_{\text{recon}}\) to \( 4 \times 10^{-5} \) at iteration 1800 and to \( 6 \times 10^{-5} \) at iteration 2000 effectively stabilizes training and prevents collapse. This suggests that the reconstruction loss serves as a regularizer for iGCT. Additionally, we observed diminishing returns when training exceeded 240k iterations, leading us to stop at 260k iterations for our experiments. These findings indicate that alternative training strategies, such as framing iGCT as a multi-task learning problem \cite{kendall2018multi,liu2019loss}, and conducting a more sophisticated analysis of loss weighting, may be necessary to enhance stability and improve convergence. See Table \ref{tab:igct_training_configs} for a summary of the training configurations for iGCT.

\begin{table}[t]
\caption{Comparison of GPU hours across the methods used in our experiments on CIFAR-10.}
\centering
\begin{tabular}{|l|c|}
\hline
\textbf{Methods} & \textbf{A100 (40G) GPU hours} \\ \hline
CFG-EDM \cite{karras2022elucidating} & 312 \\ \hline
Guided-CD \cite{song2023consistency} & 3968 \\ \hline
iGCT (ours) & 2032 \\ \hline
\end{tabular}
\label{table:compute_resources}
\end{table}

\begin{figure*}[t!]  
    \centering
    \begin{subfigure}[b]{0.33\textwidth}
    \includegraphics[width=\textwidth]{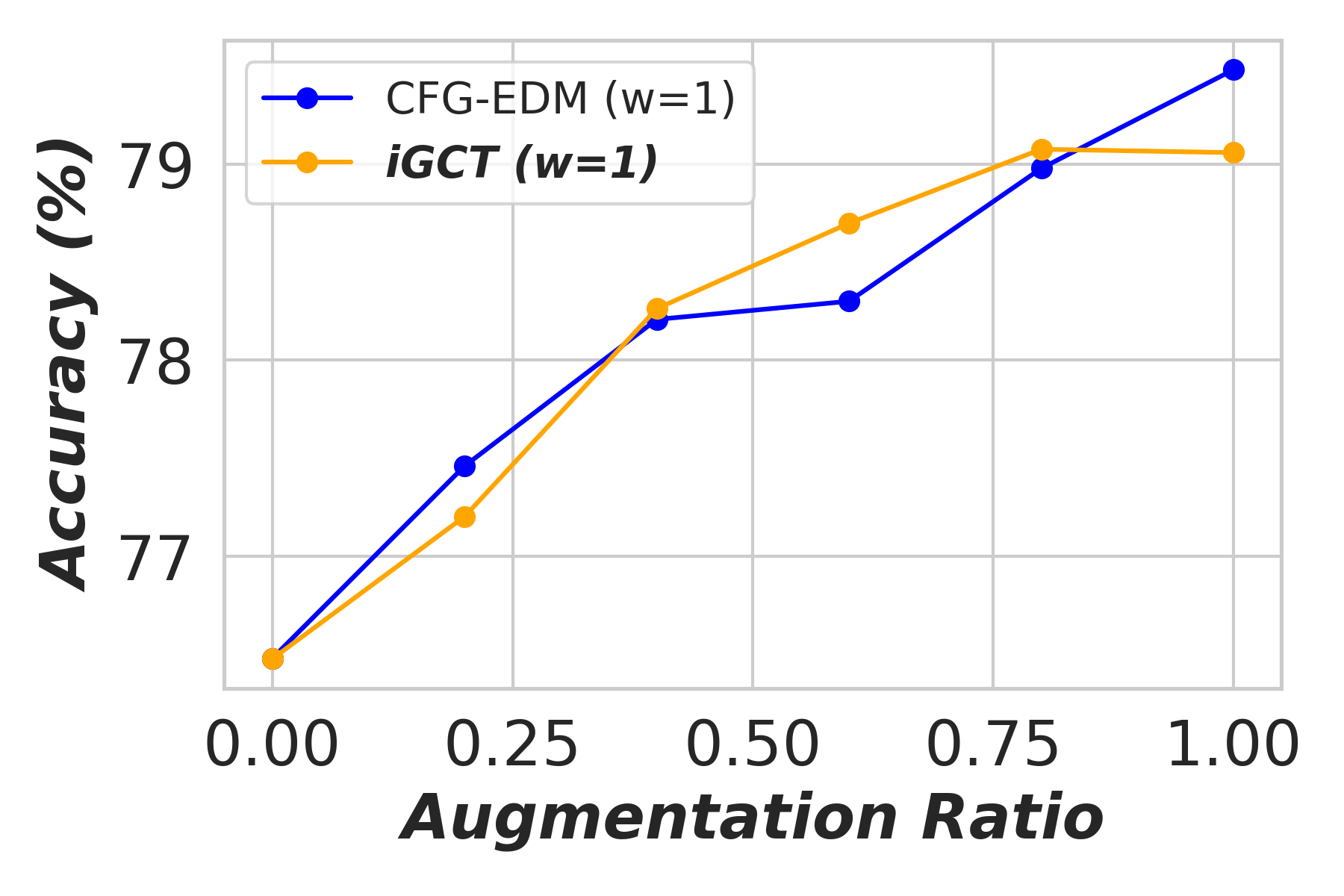} 
        \caption{Accuracy on various ratios of augmented data, guidance scale w=1.}
    \end{subfigure}
    \begin{subfigure}[b]{0.33\textwidth}
    \includegraphics[width=\textwidth]{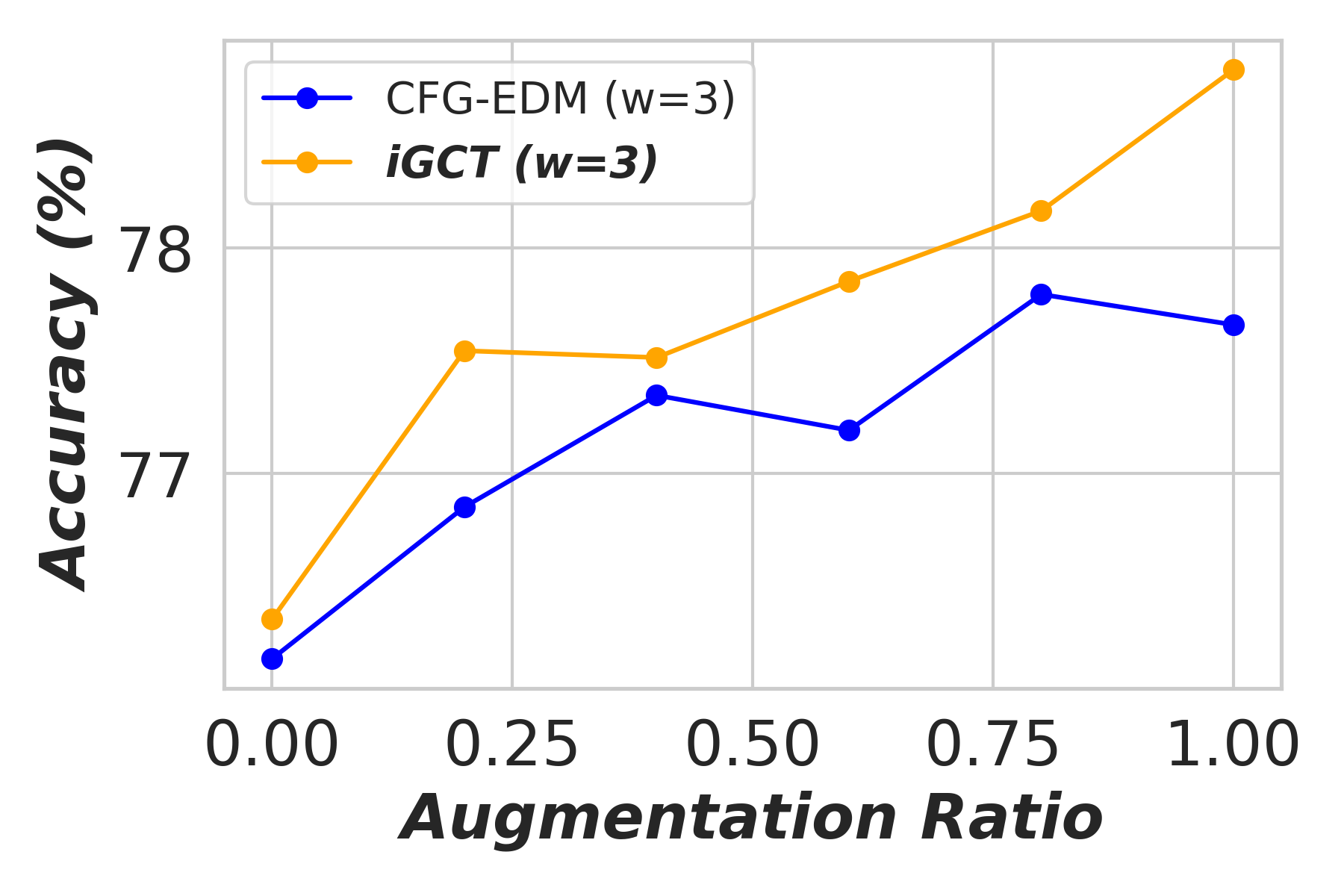} 
        \caption{Accuracy on various ratios of augmented data, guidance scale w=3.}
    \end{subfigure}
    \begin{subfigure}[b]{0.33\textwidth}
    \includegraphics[width=\textwidth]{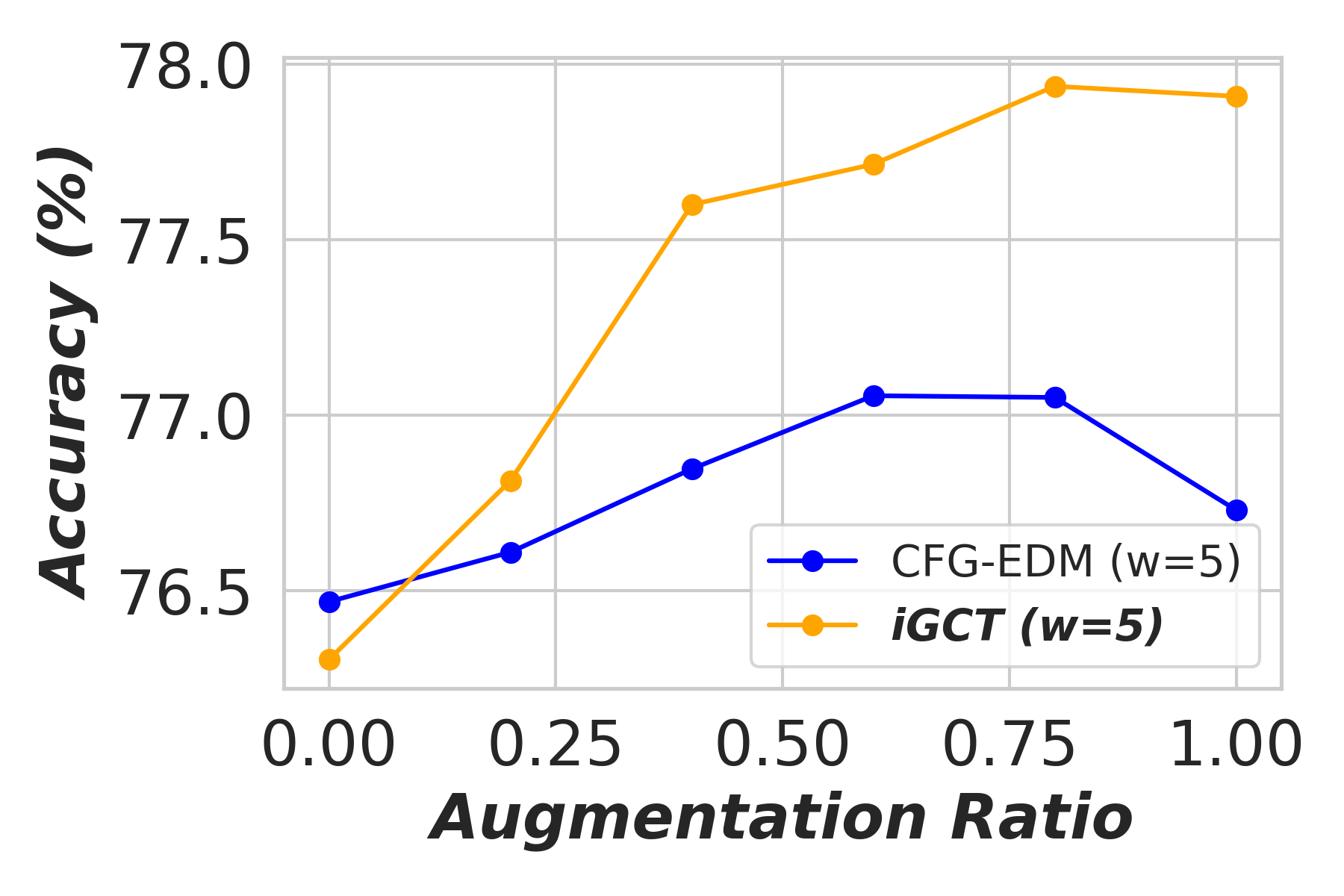} 
        \caption{Accuracy on various ratios of augmented data, guidance scale w=5.}
    \end{subfigure}
    \begin{subfigure}[b]{0.33\textwidth}
    \includegraphics[width=\textwidth]{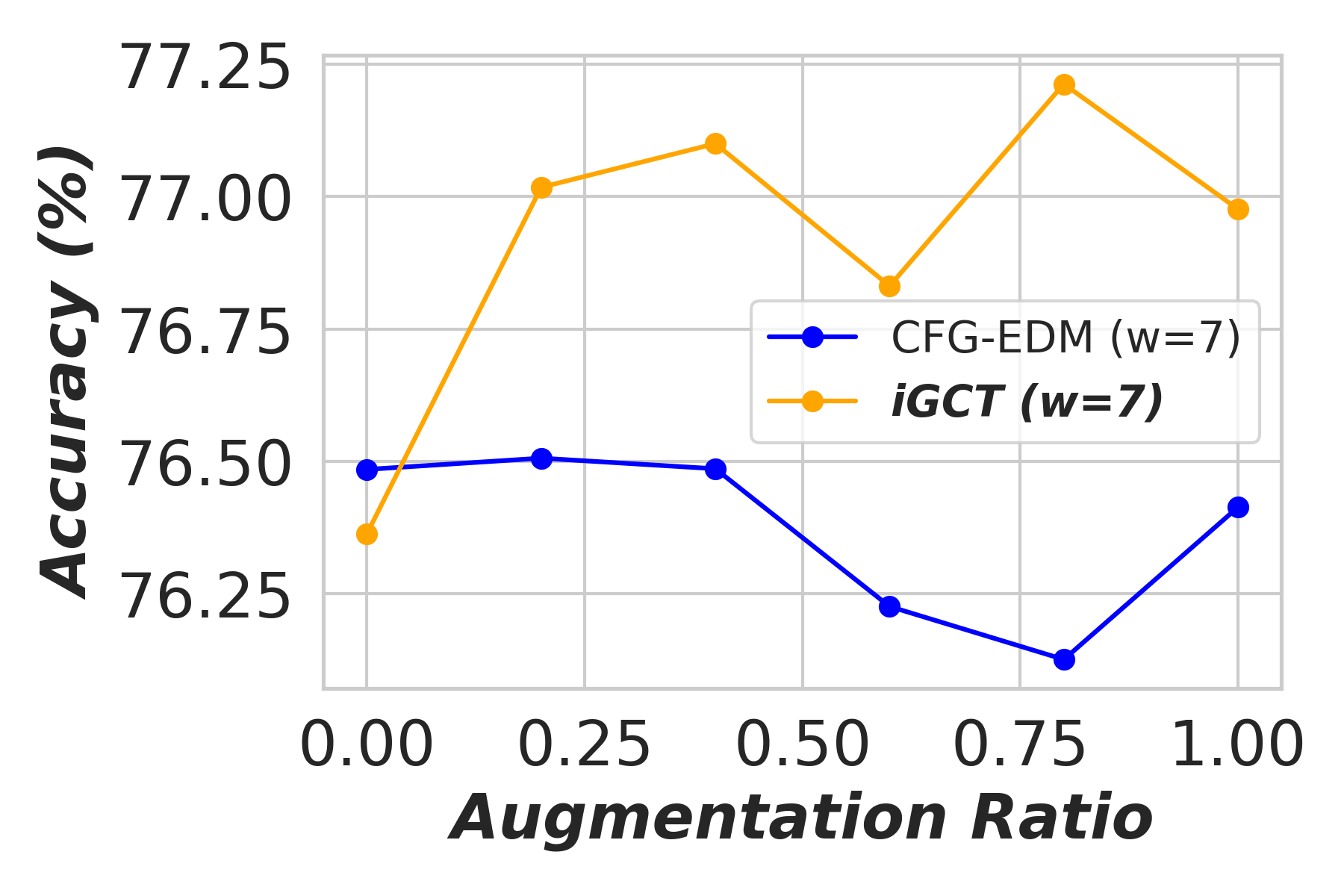} 
        \caption{Accuracy on various ratios of augmented data, guidance scale w=7.}
    \end{subfigure}
    \begin{subfigure}[b]{0.33\textwidth}
    \includegraphics[width=\textwidth]{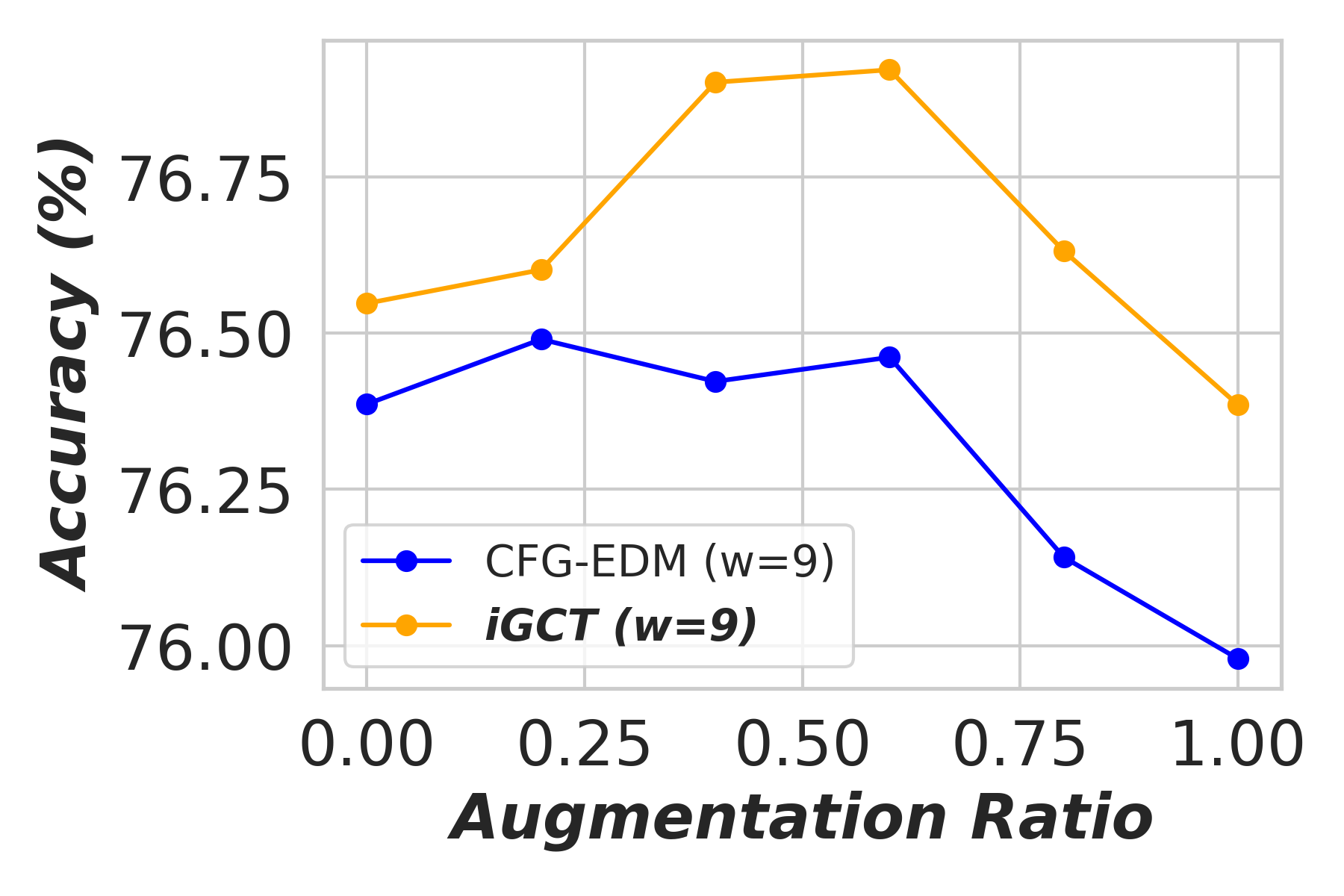} 
        \caption{Accuracy on various ratios of augmented data, guidance scale w=9.}
    \end{subfigure}
    \caption{Comparison of synthesized methods, CFG-EDM vs iGCT, used for data augmentation in image classification. iGCT consistently improves accuracy. Conversely, augmentation data synthesized from CFG-EDM offers only limited gains.}
    \vspace{-1.5em}
    \label{fig:cls_results}
\end{figure*}

\vspace{-0.1cm}
\section{Application: Data Augmentation Under Different Guidance}
\vspace{-0.2cm}

In this section, we show the effectiveness of data augmentation with diffusion-based models, CFG-EDM and iGCT, across varying guidance scales for image classification on CIFAR-10 \cite{article}. High quality data augmentation has been shown to enhance classification performance \cite{yang2023imagedataaugmentationdeep}. Under high guidance, augmentation data generated from iGCT consistently improves accuracy. Conversely, augmentation data synthesized from CFG-EDM offers only limited gains. We describe the ratios of real to synthesized data, the classifier architecture, and the training setup in the following. 

\noindent{\bf Training Details.} We conduct classification experiments trained on six different mixtures of augmented data synthesized by iGCT and CFG-EDM: \(0\%\), \(20\%\), \(40\%\), \(80\%\), and \(100\%\). The ratio represents \(\textit{synthesized data} / \textit{real data}\). For example, \(0\%\) indicates that the training and validation sets contain only 50k of real samples from CIFAR-10, and \(20\%\) includes 50k real \textit{and} 10k synthesized samples. In terms of guidance scales, we choose \(w=1,3,5,7,9\) to synthesize the augmented data using iGCT and CFG-EDM. 
The augmented dataset is split 80/20 for training and validation. For testing, the model is evaluated on the CIFAR-10 test set with 10k samples and ground truth labels. 

The standard ResNet-18 \cite{he2015deepresiduallearningimage} is used to train on all different augmented datasets. All models are trained for 250 epochs, with batch size 64, using an Adam optimizer \cite{kingma2017adammethodstochasticoptimization}. For each augmentation dataset, we train the model six times under different seeds and report the average classification accuracy.

\noindent{\bf Results.} The classifier's accuracy, trained on augmented data synthesized by CFG-EDM and iGCT, is shown in Fig. \ref{fig:cls_results}. With \(w=1\) (no guidance), both iGCT and CFG-EDM provide comparable performance boosts. As guidance scale increases, iGCT shows more significant improvements than CFG-EDM. At high guidance and augmentation ratios, performance drops, but this effect occurs later for iGCT (e.g., at \(100\%\) augmentation and \(w=9\)), while CFG-EDM stops improving accuracy at \(w=7\). This experiment highlights the importance of high-quality data under high guidance, with iGCT outperforming CFG-EDM in data quality.

\section{Uncurated Results}
In this section, we present additional qualitative results to highlight the performance of our proposed iGCT method compared to the multi-step EDM baseline. These visualizations include both inversion and guidance tasks across the CIFAR-10 and ImageNet64 datasets. The results demonstrate iGCT's ability to maintain competitive quality with significantly fewer steps and minimal artifacts, showcasing the effectiveness of our approach.

\subsection{Inversion Results}
We provide additional visualization of the latent noise on both CIFAR-10 and ImageNet64 datasets. Fig. \ref{fig:CIFAR-10_inversion_reconstruction} and Fig. \ref{fig:im64_inversion_reconstruction} compare our 1-step iGCT with the multi-step EDM on inversion and reconstruction.  

\subsection{Editing Results}
\begin{figure*}[t]
    \centering

    \begin{minipage}{0.48\textwidth}
        \centering
        \begin{subfigure}{0.48\textwidth}
            \includegraphics[width=\linewidth]{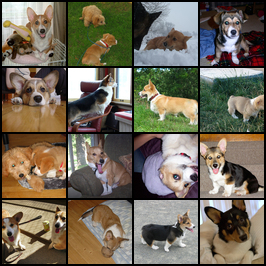}
            \caption{Original: "corgi"}
        \end{subfigure}

        \begin{subfigure}{0.48\textwidth}
            \includegraphics[width=\linewidth]{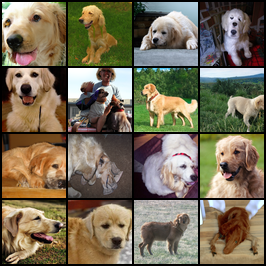}
            \caption{EDM (18 NFE), w=1}
        \end{subfigure}
        \begin{subfigure}{0.48\textwidth}
            \includegraphics[width=\linewidth]{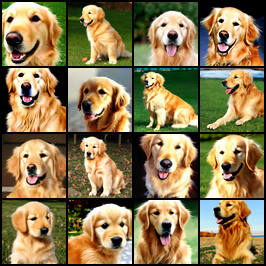}
            \caption{EDM (18 NFE), w=7}
        \end{subfigure}
        \begin{subfigure}{0.48\textwidth}
            \includegraphics[width=\linewidth]{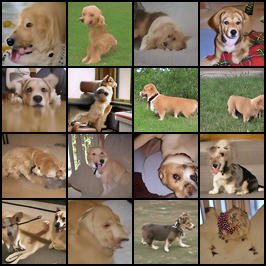}
            \caption{iGCT (1 NFE), w=7}
        \end{subfigure}
        \begin{subfigure}{0.48\textwidth}
            \includegraphics[width=\linewidth]{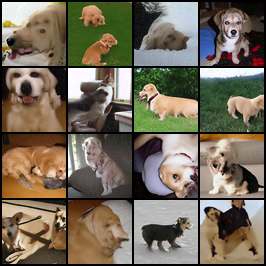}
            \caption{iGCT (1 NFE), w=1}
        \end{subfigure}

        \caption{ImageNet64: "corgi" $\rightarrow$ "golden retriever"}
        \label{fig:im64_edit_1}
    \end{minipage}
    \hfill
    \begin{minipage}{0.48\textwidth}
        \centering
        \begin{subfigure}{0.48\textwidth}
            \includegraphics[width=\linewidth]{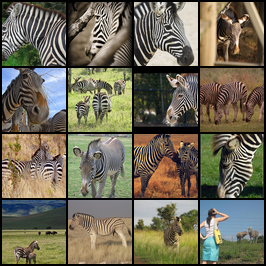}
            \caption{Original: "zebra"}
        \end{subfigure}

        \begin{subfigure}{0.48\textwidth}
            \includegraphics[width=\linewidth]{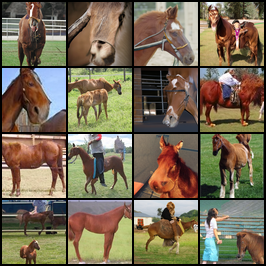}
            \caption{EDM (18 NFE), w=1}
        \end{subfigure}
        \begin{subfigure}{0.48\textwidth}
            \includegraphics[width=\linewidth]{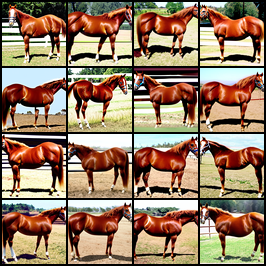}
            \caption{EDM (18 NFE), w=7}
        \end{subfigure}
        \begin{subfigure}{0.48\textwidth}
            \includegraphics[width=\linewidth]{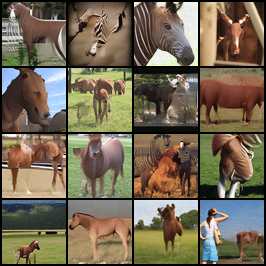}
            \caption{iGCT (1 NFE), w=1}
        \end{subfigure}
        \begin{subfigure}{0.48\textwidth}
            \includegraphics[width=\linewidth]{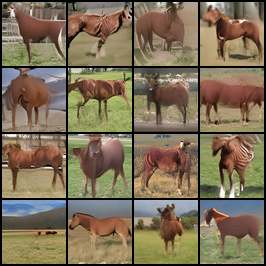}
            \caption{iGCT (1 NFE), w=7}
        \end{subfigure}

        \caption{ImageNet64: "zebra" $\rightarrow$ "horse"}
        \label{fig:im64_edit_2}
    \end{minipage}

\end{figure*}

\begin{figure*}[t]
    \centering

    \begin{minipage}{0.48\textwidth}
        \centering
        \begin{subfigure}{0.48\textwidth}
            \includegraphics[width=\linewidth]{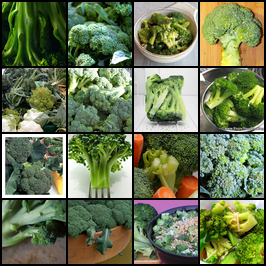}
            \caption{Original: "broccoli"}
        \end{subfigure}

        \begin{subfigure}{0.48\textwidth}
            \includegraphics[width=\linewidth]{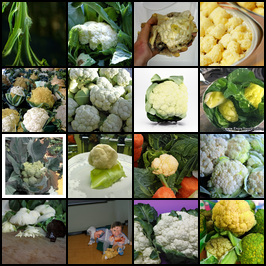}
            \caption{EDM (18 NFE), w=1}
        \end{subfigure}
        \begin{subfigure}{0.48\textwidth}
            \includegraphics[width=\linewidth]{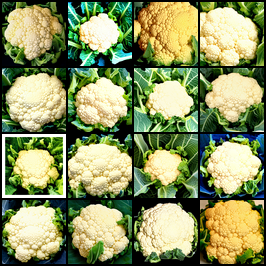}
            \caption{EDM (18 NFE), w=7}
        \end{subfigure}
        \begin{subfigure}{0.48\textwidth}
            \includegraphics[width=\linewidth]{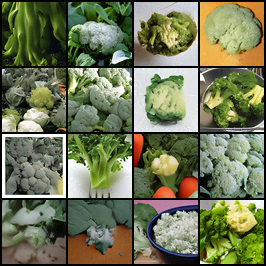}
            \caption{iGCT (1 NFE), w=1}
        \end{subfigure}
        \begin{subfigure}{0.48\textwidth}
            \includegraphics[width=\linewidth]{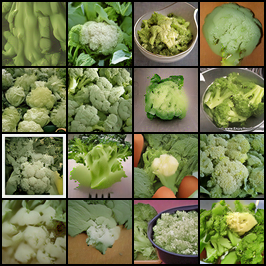}
            \caption{iGCT (1 NFE), w=7}
        \end{subfigure}

        \caption{ImageNet64: "broccoli" $\rightarrow$ "cauliflower"}
        \label{fig:im64_edit_3}
    \end{minipage}
    \hfill
    \begin{minipage}{0.48\textwidth}
        \centering
        \begin{subfigure}{0.48\textwidth}
            \includegraphics[width=\linewidth]{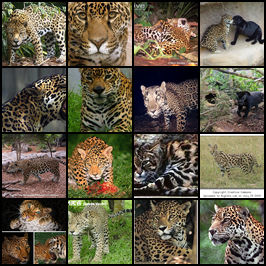}
            \caption{Original: "jaguar"}
        \end{subfigure}

        \begin{subfigure}{0.48\textwidth}
            \includegraphics[width=\linewidth]{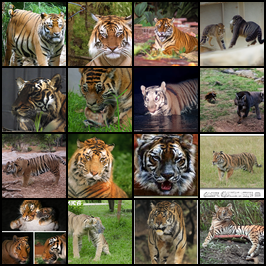}
            \caption{EDM (18 NFE), w=1}
        \end{subfigure}
        \begin{subfigure}{0.48\textwidth}
            \includegraphics[width=\linewidth]{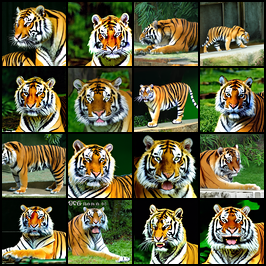}
            \caption{EDM (18 NFE), w=7}
        \end{subfigure}
        \begin{subfigure}{0.48\textwidth}
            \includegraphics[width=\linewidth]{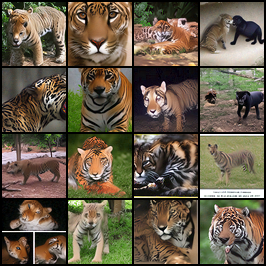}
            \caption{iGCT (1 NFE), w=1}
        \end{subfigure}
        \begin{subfigure}{0.48\textwidth}
            \includegraphics[width=\linewidth]{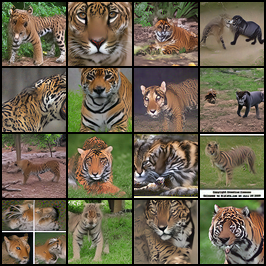}
            \caption{iGCT (1 NFE), w=7}
        \end{subfigure}

        \caption{ImageNet64: "jaguar" $\rightarrow$ "tiger"}
        \label{fig:im64_edit_4}
    \end{minipage}

\end{figure*}

In this section, we dump more uncurated editing results on ImageNet64's subgroups mentioned in Sec. \ref{sec:image-editing}. Fig. \ref{fig:im64_edit_1}--\ref{fig:im64_edit_4} illustrate a comparison between our 1-step iGCT and the multi-step EDM approach.

\subsection{Guidance Results}
\begin{figure*}[b]
    \centering
    \begin{subfigure}{0.25\textwidth}
        \includegraphics[width=\linewidth]{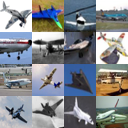}
        \caption{CFG-EDM (18 NFE), w=1.0}
    \end{subfigure}
    \begin{subfigure}{0.25\textwidth}
        \includegraphics[width=\linewidth]{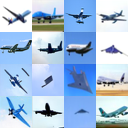}
        \caption{CFG-EDM (18 NFE), w=7.0}
    \end{subfigure}
    \begin{subfigure}{0.25\textwidth}
        \includegraphics[width=\linewidth]{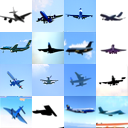}
        \caption{CFG-EDM (18 NFE), w=13.0}
    \end{subfigure}
    \begin{subfigure}{0.25\textwidth}
        \includegraphics[width=\linewidth]{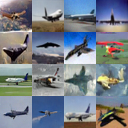}
        \caption{iGCT (1 NFE), w=1.0}
    \end{subfigure}
    \begin{subfigure}{0.25\textwidth}
        \includegraphics[width=\linewidth]{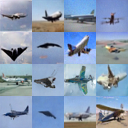}
        \caption{iGCT (1 NFE), w=7.0}
    \end{subfigure}
    \begin{subfigure}{0.25\textwidth}
        \includegraphics[width=\linewidth]{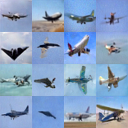}
        \caption{iGCT (1 NFE), w=13.0}
    \end{subfigure}
    \caption{CIFAR-10 "airplane"}
    \label{fig:CIFAR-10_guided_1}
\end{figure*}

\begin{figure*}[t]
    \centering
    \begin{subfigure}{0.25\textwidth}
        \includegraphics[width=\linewidth]{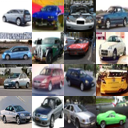}
        \caption{CFG-EDM (18 NFE), w=1.0}
    \end{subfigure}
    \begin{subfigure}{0.25\textwidth}
        \includegraphics[width=\linewidth]{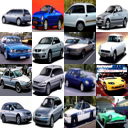}
        \caption{CFG-EDM (18 NFE), w=7.0}
    \end{subfigure}
    \begin{subfigure}{0.25\textwidth}
        \includegraphics[width=\linewidth]{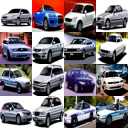}
        \caption{CFG-EDM (18 NFE), w=13.0}
    \end{subfigure}
    \begin{subfigure}{0.25\textwidth}
        \includegraphics[width=\linewidth]{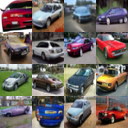}
        \caption{iGCT (1 NFE), w=1.0}
    \end{subfigure}
    \begin{subfigure}{0.25\textwidth}
        \includegraphics[width=\linewidth]{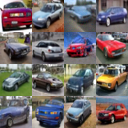}
        \caption{iGCT (1 NFE), w=7.0}
    \end{subfigure}
    \begin{subfigure}{0.25\textwidth}
        \includegraphics[width=\linewidth]{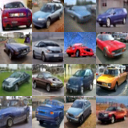}
        \caption{iGCT (1 NFE), w=13.0}
    \end{subfigure}
    \caption{CIFAR-10 "car"}
    \label{fig:CIFAR-10_guided_2}
\end{figure*}

\begin{figure*}[t]
    \centering
    \begin{subfigure}{0.25\textwidth}
        \includegraphics[width=\linewidth]{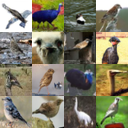}
        \caption{CFG-EDM (18 NFE), w=1.0}
    \end{subfigure}
    \begin{subfigure}{0.25\textwidth}
        \includegraphics[width=\linewidth]{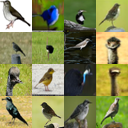}
        \caption{CFG-EDM (18 NFE), w=7.0}
    \end{subfigure}
    \begin{subfigure}{0.25\textwidth}
        \includegraphics[width=\linewidth]{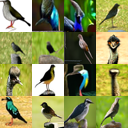}
        \caption{CFG-EDM (18 NFE), w=13.0}
    \end{subfigure}
    \begin{subfigure}{0.25\textwidth}
        \includegraphics[width=\linewidth]{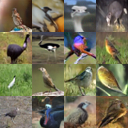}
        \caption{iGCT (1 NFE), w=1.0}
    \end{subfigure}
    \begin{subfigure}{0.25\textwidth}
        \includegraphics[width=\linewidth]{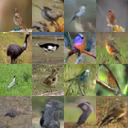}
        \caption{iGCT (1 NFE), w=7.0}
    \end{subfigure}
    \begin{subfigure}{0.25\textwidth}
        \includegraphics[width=\linewidth]{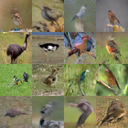}
        \caption{iGCT (1 NFE), w=13.0}
    \end{subfigure}
    \caption{CIFAR-10 "bird"}
    \label{fig:CIFAR-10_guided_3}
\end{figure*}

\begin{figure*}[t]
    \centering
    \begin{subfigure}{0.25\textwidth}
        \includegraphics[width=\linewidth]{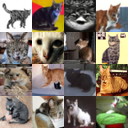}
        \caption{CFG-EDM (18 NFE), w=1.0}
    \end{subfigure}
    \begin{subfigure}{0.25\textwidth}
        \includegraphics[width=\linewidth]{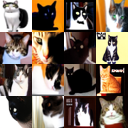}
        \caption{CFG-EDM (18 NFE), w=7.0}
    \end{subfigure}
    \begin{subfigure}{0.25\textwidth}
        \includegraphics[width=\linewidth]{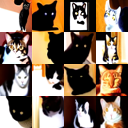}
        \caption{CFG-EDM (18 NFE), w=13.0}
    \end{subfigure}
    \begin{subfigure}{0.25\textwidth}
        \includegraphics[width=\linewidth]{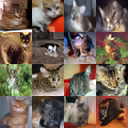}
        \caption{iGCT (1 NFE), w=1.0}
    \end{subfigure}
    \begin{subfigure}{0.25\textwidth}
        \includegraphics[width=\linewidth]{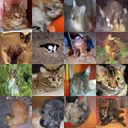}
        \caption{iGCT (1 NFE), w=7.0}
    \end{subfigure}
    \begin{subfigure}{0.25\textwidth}
        \includegraphics[width=\linewidth]{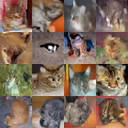}
        \caption{iGCT (1 NFE), w=13.0}
    \end{subfigure}
    \caption{CIFAR-10 "cat"}
    \label{fig:CIFAR-10_guided_4}
\end{figure*}

\begin{figure*}[t]
    \centering
    \begin{subfigure}{0.25\textwidth}
        \includegraphics[width=\linewidth]{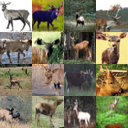}
        \caption{CFG-EDM (18 NFE), w=1.0}
    \end{subfigure}
    \begin{subfigure}{0.25\textwidth}
        \includegraphics[width=\linewidth]{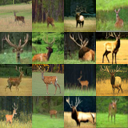}
        \caption{CFG-EDM (18 NFE), w=7.0}
    \end{subfigure}
    \begin{subfigure}{0.25\textwidth}
        \includegraphics[width=\linewidth]{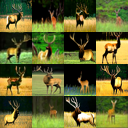}
        \caption{CFG-EDM (18 NFE), w=13.0}
    \end{subfigure}
    \begin{subfigure}{0.25\textwidth}
        \includegraphics[width=\linewidth]{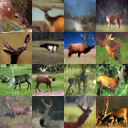}
        \caption{iGCT (1 NFE), w=1.0}
    \end{subfigure}
    \begin{subfigure}{0.25\textwidth}
        \includegraphics[width=\linewidth]{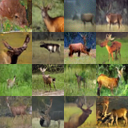}
        \caption{iGCT (1 NFE), w=7.0}
    \end{subfigure}
    \begin{subfigure}{0.25\textwidth}
        \includegraphics[width=\linewidth]{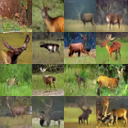}
        \caption{iGCT (1 NFE), w=13.0}
    \end{subfigure}
    \caption{CIFAR-10 "deer"}
    \label{fig:CIFAR-10_guided_5}
\end{figure*}

\begin{figure*}[t]
    \centering
    \begin{subfigure}{0.25\textwidth}
        \includegraphics[width=\linewidth]{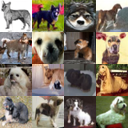}
        \caption{CFG-EDM (18 NFE), w=1.0}
    \end{subfigure}
    \begin{subfigure}{0.25\textwidth}
        \includegraphics[width=\linewidth]{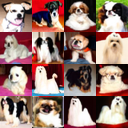}
        \caption{CFG-EDM (18 NFE), w=7.0}
    \end{subfigure}
    \begin{subfigure}{0.25\textwidth}
        \includegraphics[width=\linewidth]{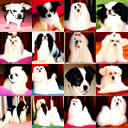}
        \caption{CFG-EDM (18 NFE), w=13.0}
    \end{subfigure}
    \begin{subfigure}{0.25\textwidth}
        \includegraphics[width=\linewidth]{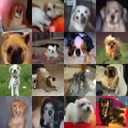}
        \caption{iGCT (1 NFE), w=1.0}
    \end{subfigure}
    \begin{subfigure}{0.25\textwidth}
        \includegraphics[width=\linewidth]{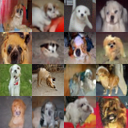}
        \caption{iGCT (1 NFE), w=7.0}
    \end{subfigure}
    \begin{subfigure}{0.25\textwidth}
        \includegraphics[width=\linewidth]{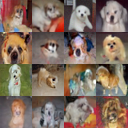}
        \caption{iGCT (1 NFE), w=13.0}
    \end{subfigure}
    \caption{CIFAR-10 "dog"}
    \label{fig:CIFAR-10_guided_6}
\end{figure*}

\begin{figure*}[t]
    \centering
    \begin{subfigure}{0.25\textwidth}
        \includegraphics[width=\linewidth]{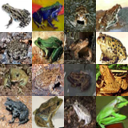}
        \caption{CFG-EDM (18 NFE), w=1.0}
    \end{subfigure}
    \begin{subfigure}{0.25\textwidth}
        \includegraphics[width=\linewidth]{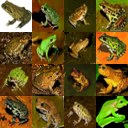}
        \caption{CFG-EDM (18 NFE), w=7.0}
    \end{subfigure}
    \begin{subfigure}{0.25\textwidth}
        \includegraphics[width=\linewidth]{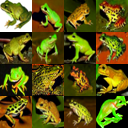}
        \caption{CFG-EDM (18 NFE), w=13.0}
    \end{subfigure}
    \begin{subfigure}{0.25\textwidth}
        \includegraphics[width=\linewidth]{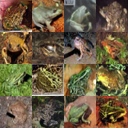}
        \caption{iGCT (1 NFE), w=1.0}
    \end{subfigure}
    \begin{subfigure}{0.25\textwidth}
        \includegraphics[width=\linewidth]{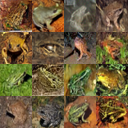}
        \caption{iGCT (1 NFE), w=7.0}
    \end{subfigure}
    \begin{subfigure}{0.25\textwidth}
        \includegraphics[width=\linewidth]{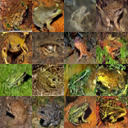}
        \caption{iGCT (1 NFE), w=13.0}
    \end{subfigure}
    \caption{CIFAR-10 "frog"}
    \label{fig:CIFAR-10_guided_7}
\end{figure*}

\begin{figure*}[t]
    \centering
    \begin{subfigure}{0.25\textwidth}
        \includegraphics[width=\linewidth]{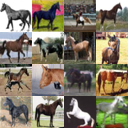}
        \caption{CFG-EDM (18 NFE), w=1.0}
    \end{subfigure}
    \begin{subfigure}{0.25\textwidth}
        \includegraphics[width=\linewidth]{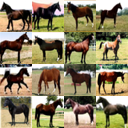}
        \caption{CFG-EDM (18 NFE), w=7.0}
    \end{subfigure}
    \begin{subfigure}{0.25\textwidth}
        \includegraphics[width=\linewidth]{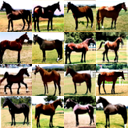}
        \caption{CFG-EDM (18 NFE), w=13.0}
    \end{subfigure}
    \begin{subfigure}{0.25\textwidth}
        \includegraphics[width=\linewidth]{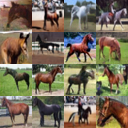}
        \caption{iGCT (1 NFE), w=1.0}
    \end{subfigure}
    \begin{subfigure}{0.25\textwidth}
        \includegraphics[width=\linewidth]{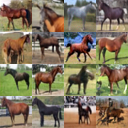}
        \caption{iGCT (1 NFE), w=7.0}
    \end{subfigure}
    \begin{subfigure}{0.25\textwidth}
        \includegraphics[width=\linewidth]{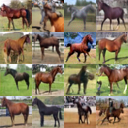}
        \caption{iGCT (1 NFE), w=13.0}
    \end{subfigure}
    \caption{CIFAR-10 "horse"}
    \label{fig:CIFAR-10_guided_8}
\end{figure*}

\begin{figure*}[t]
    \centering
    \begin{subfigure}{0.25\textwidth}
        \includegraphics[width=\linewidth]{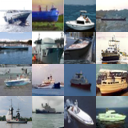}
        \caption{CFG-EDM (18 NFE), w=1.0}
    \end{subfigure}
    \begin{subfigure}{0.25\textwidth}
        \includegraphics[width=\linewidth]{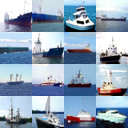}
        \caption{CFG-EDM (18 NFE), w=7.0}
    \end{subfigure}
    \begin{subfigure}{0.25\textwidth}
        \includegraphics[width=\linewidth]{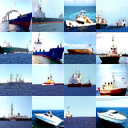}
        \caption{CFG-EDM (18 NFE), w=13.0}
    \end{subfigure}
    \begin{subfigure}{0.25\textwidth}
        \includegraphics[width=\linewidth]{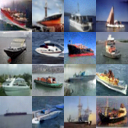}
        \caption{iGCT (1 NFE), w=1.0}
    \end{subfigure}
    \begin{subfigure}{0.25\textwidth}
        \includegraphics[width=\linewidth]{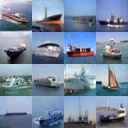}
        \caption{iGCT (1 NFE), w=7.0}
    \end{subfigure}
    \begin{subfigure}{0.25\textwidth}
        \includegraphics[width=\linewidth]{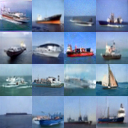}
        \caption{iGCT (1 NFE), w=13.0}
    \end{subfigure}
    \caption{CIFAR-10 "ship"}
    \label{fig:CIFAR-10_guided_9}
\end{figure*}

\begin{figure*}[t]
    \centering
    \begin{subfigure}{0.25\textwidth}
        \includegraphics[width=\linewidth]{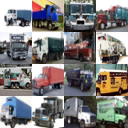}
        \caption{CFG-EDM (18 NFE), w=1.0}
    \end{subfigure}
    \begin{subfigure}{0.25\textwidth}
        \includegraphics[width=\linewidth]{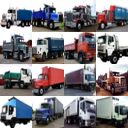}
        \caption{CFG-EDM (18 NFE), w=7.0}
    \end{subfigure}
    \begin{subfigure}{0.25\textwidth}
        \includegraphics[width=\linewidth]{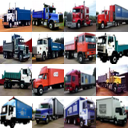}
        \caption{CFG-EDM (18 NFE), w=13.0}
    \end{subfigure}
    \begin{subfigure}{0.25\textwidth}
        \includegraphics[width=\linewidth]{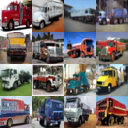}
        \caption{iGCT (1 NFE), w=1.0}
    \end{subfigure}
    \begin{subfigure}{0.25\textwidth}
        \includegraphics[width=\linewidth]{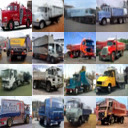}
        \caption{iGCT (1 NFE), w=7.0}
    \end{subfigure}
    \begin{subfigure}{0.25\textwidth}
        \includegraphics[width=\linewidth]{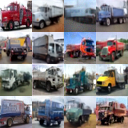}
        \caption{iGCT (1 NFE), w=13.0}
    \end{subfigure}
    \caption{CIFAR-10 "truck"}
    \label{fig:CIFAR-10_guided_10}
\end{figure*}

\begin{figure*}[b]
    \centering
    \begin{subfigure}{0.25\textwidth}
        \includegraphics[width=\linewidth]{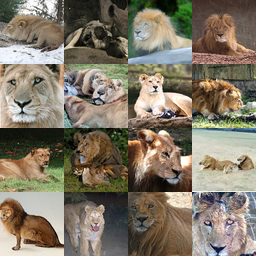}
        \caption{CFG-EDM (18 NFE), w=1.0}
    \end{subfigure}
    \begin{subfigure}{0.25\textwidth}
        \includegraphics[width=\linewidth]{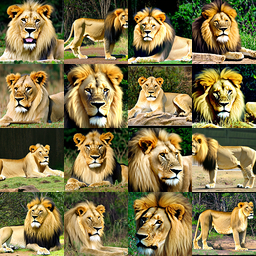}
        \caption{CFG-EDM (18 NFE), w=7.0}
    \end{subfigure}
    \begin{subfigure}{0.25\textwidth}
        \includegraphics[width=\linewidth]{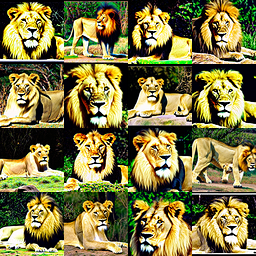}
        \caption{CFG-EDM (18 NFE), w=13.0}
    \end{subfigure}
    \begin{subfigure}{0.25\textwidth}
        \includegraphics[width=\linewidth]{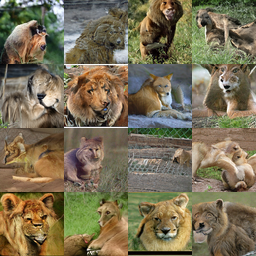}
        \caption{iGCT (2 NFE), w=1.0}
    \end{subfigure}
    \begin{subfigure}{0.25\textwidth}
        \includegraphics[width=\linewidth]{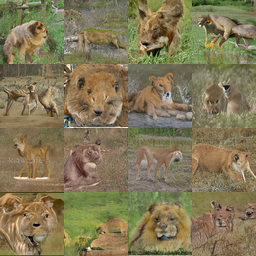}
        \caption{iGCT (2 NFE), w=7.0}
    \end{subfigure}
    \begin{subfigure}{0.25\textwidth}
        \includegraphics[width=\linewidth]{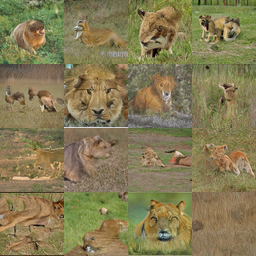}
        \caption{iGCT (2 NFE), w=13.0}
    \end{subfigure}
    \caption{ImageNet64 "lion"}
    \label{fig:im64_guided_1}
\end{figure*}

\begin{figure*}[b]
    \centering
    \begin{subfigure}{0.25\textwidth}
        \includegraphics[width=\linewidth]{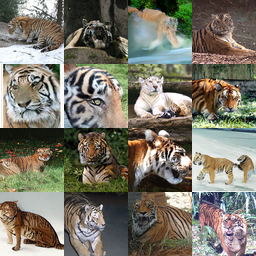}
        \caption{CFG-EDM (18 NFE), w=1.0}
    \end{subfigure}
    \begin{subfigure}{0.25\textwidth}
        \includegraphics[width=\linewidth]{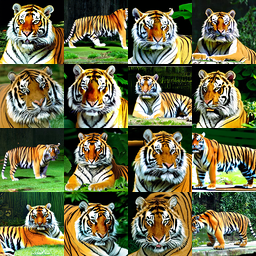}
        \caption{CFG-EDM (18 NFE), w=7.0}
    \end{subfigure}
    \begin{subfigure}{0.25\textwidth}
        \includegraphics[width=\linewidth]{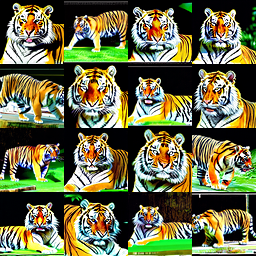}
        \caption{CFG-EDM (18 NFE), w=13.0}
    \end{subfigure}
    \begin{subfigure}{0.25\textwidth}
        \includegraphics[width=\linewidth]{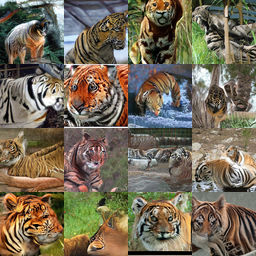}
        \caption{iGCT (2 NFE), w=1.0}
    \end{subfigure}
    \begin{subfigure}{0.25\textwidth}
        \includegraphics[width=\linewidth]{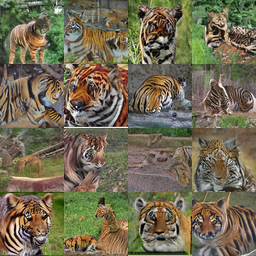}
        \caption{iGCT (2 NFE), w=7.0}
    \end{subfigure}
    \begin{subfigure}{0.25\textwidth}
        \includegraphics[width=\linewidth]{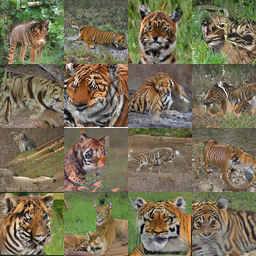}
        \caption{iGCT (2 NFE), w=13.0}
    \end{subfigure}
    \caption{ImageNet64 "tiger"}
    \label{fig:im64_guided_2}
\end{figure*}

\begin{figure*}[b]
    \centering
    \begin{subfigure}{0.25\textwidth}
        \includegraphics[width=\linewidth]{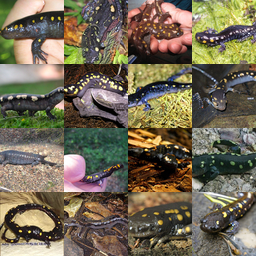}
        \caption{CFG-EDM (18 NFE), w=1.0}
    \end{subfigure}
    \begin{subfigure}{0.25\textwidth}
        \includegraphics[width=\linewidth]{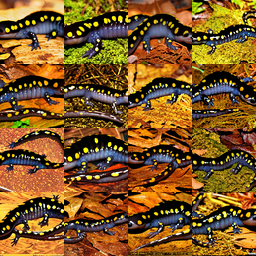}
        \caption{CFG-EDM (18 NFE), w=7.0}
    \end{subfigure}
    \begin{subfigure}{0.25\textwidth}
        \includegraphics[width=\linewidth]{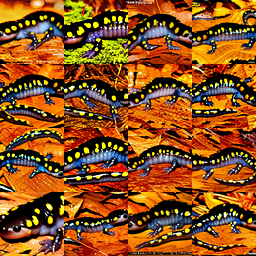}
        \caption{CFG-EDM (18 NFE), w=13.0}
    \end{subfigure}
    \begin{subfigure}{0.25\textwidth}
        \includegraphics[width=\linewidth]{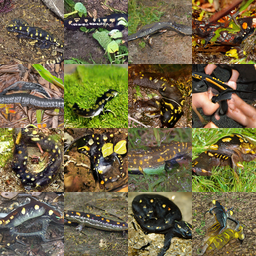}
        \caption{iGCT (2 NFE), w=1.0}
    \end{subfigure}
    \begin{subfigure}{0.25\textwidth}
        \includegraphics[width=\linewidth]{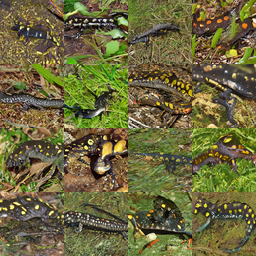}
        \caption{iGCT (2 NFE), w=7.0}
    \end{subfigure}
    \begin{subfigure}{0.25\textwidth}
        \includegraphics[width=\linewidth]{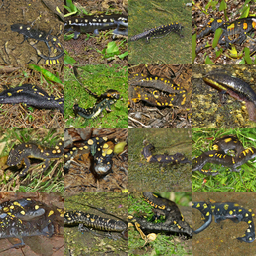}
        \caption{iGCT (2 NFE), w=13.0}
    \end{subfigure}
    \caption{ImageNet64 "salamander"}
    \label{fig:im64_guided_3}
\end{figure*}

\begin{figure*}[b]
    \centering
    \begin{subfigure}{0.25\textwidth}
        \includegraphics[width=\linewidth]{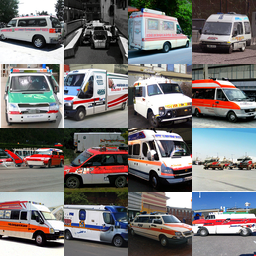}
        \caption{CFG-EDM (18 NFE), w=1.0}
    \end{subfigure}
    \begin{subfigure}{0.25\textwidth}
        \includegraphics[width=\linewidth]{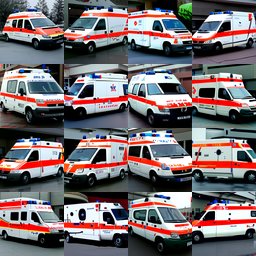}
        \caption{CFG-EDM (18 NFE), w=7.0}
    \end{subfigure}
    \begin{subfigure}{0.25\textwidth}
        \includegraphics[width=\linewidth]{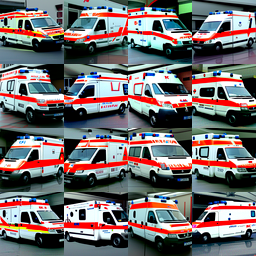}
        \caption{CFG-EDM (18 NFE), w=13.0}
    \end{subfigure}
    \begin{subfigure}{0.25\textwidth}
        \includegraphics[width=\linewidth]{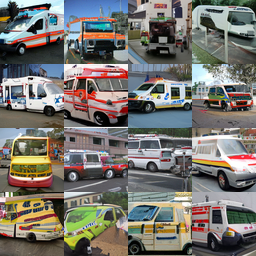}
        \caption{iGCT (2 NFE), w=1.0}
    \end{subfigure}
    \begin{subfigure}{0.25\textwidth}
        \includegraphics[width=\linewidth]{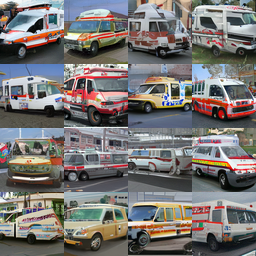}
        \caption{iGCT (2 NFE), w=7.0}
    \end{subfigure}
    \begin{subfigure}{0.25\textwidth}
        \includegraphics[width=\linewidth]{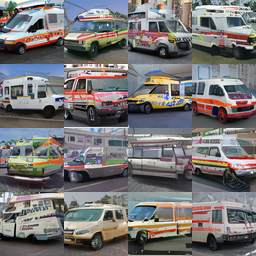}
        \caption{iGCT (2 NFE), w=13.0}
    \end{subfigure}
    \caption{ImageNet64 "ambulance"}
    \label{fig:im64_guided_4}
\end{figure*}

In Section \ref{sec:guidance}, we demonstrated that iGCT provides a guidance solution without introducing the high-contrast artifacts commonly observed in CFG-based methods. Here, we present additional uncurated results on CIFAR-10 and ImageNet64. For CIFAR-10, iGCT achieves competitive performance compared to the baseline diffusion model, which requires multiple steps for generation. See Figs. \ref{fig:CIFAR-10_guided_1}--\ref{fig:CIFAR-10_guided_10}. For ImageNet64, although the visual quality of iGCT's generated images falls slightly short of expectations, this can be attributed to the smaller UNet architecture used—only 61\% of the baseline model size—and the need for a more robust training curriculum to prevent collapse, as discussed in Section \ref{appendix:bs-config}. Nonetheless, even at higher guidance levels, iGCT maintains style consistency, whereas CFG-based methods continue to suffer from pronounced high-contrast artifacts. See Figs. \ref{fig:im64_guided_1}--\ref{fig:im64_guided_4}.

\begin{figure*}[t]
    \centering
    \begin{subfigure}{0.48\textwidth}
        \centering
        \includegraphics[width=\linewidth]{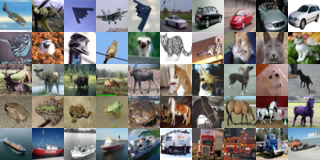}
        \caption{CIFAR-10: Original data}
    \end{subfigure}
    \begin{subfigure}{0.48\textwidth}
        \centering
        \includegraphics[width=\linewidth]{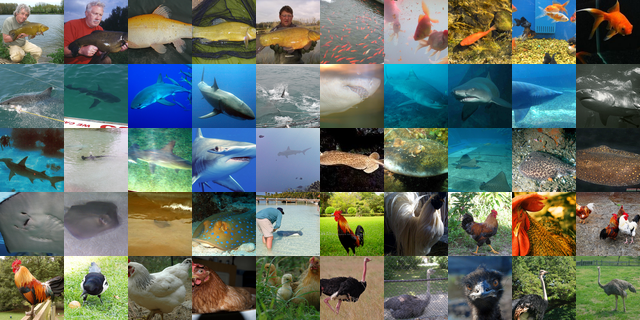}
        \caption{ImageNet64: Original data}
    \end{subfigure}

    \begin{subfigure}{0.48\textwidth}
        \centering
        \includegraphics[width=\linewidth]{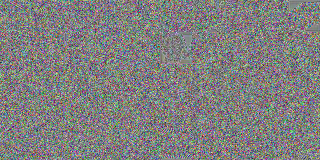}
    \end{subfigure}
    \begin{subfigure}{0.48\textwidth}
        \centering
        \includegraphics[width=\linewidth]{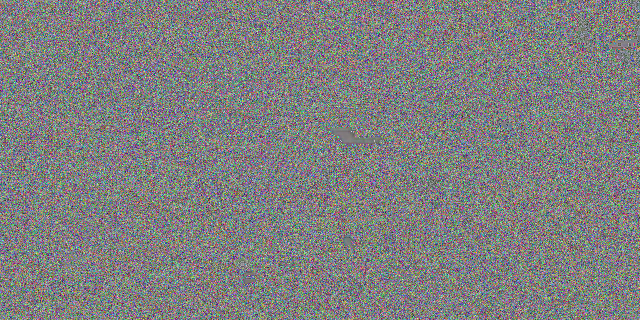}
    \end{subfigure}

    \begin{subfigure}{0.48\textwidth}
        \centering
        \includegraphics[width=\linewidth]{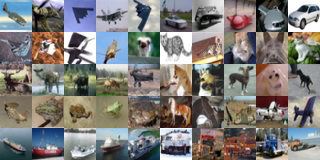}
        \caption{CIFAR-10: Inversion + reconstruction, EDM (18 NFE)}
    \end{subfigure}
    \begin{subfigure}{0.48\textwidth}
        \centering
        \includegraphics[width=\linewidth]{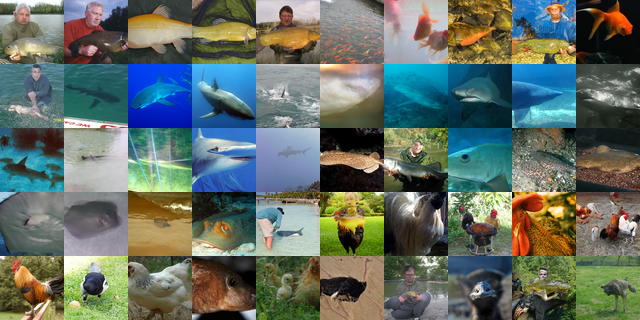}
        \caption{ImageNet64: Inversion + reconstruction, EDM (18 NFE)}
    \end{subfigure}

    \begin{subfigure}{0.48\textwidth}
        \centering
        \includegraphics[width=\linewidth]{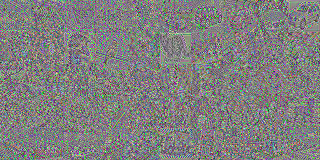}
    \end{subfigure}
    \begin{subfigure}{0.48\textwidth}
        \centering
        \includegraphics[width=\linewidth]{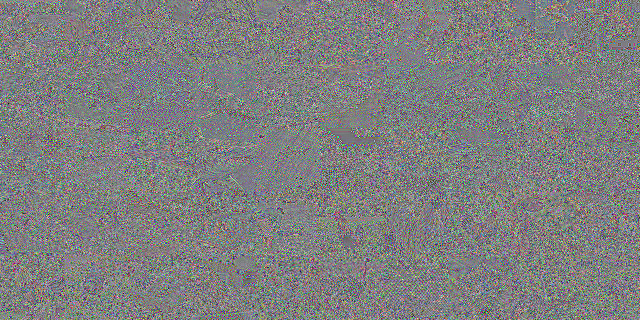}
    \end{subfigure}

    \begin{subfigure}{0.48\textwidth}
        \centering
        \includegraphics[width=\linewidth]{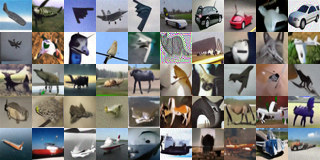}
        \caption{CIFAR-10: Inversion + reconstruction, iGCT (1 NFE)}
    \end{subfigure}
    \begin{subfigure}{0.48\textwidth}
        \centering
        \includegraphics[width=\linewidth]{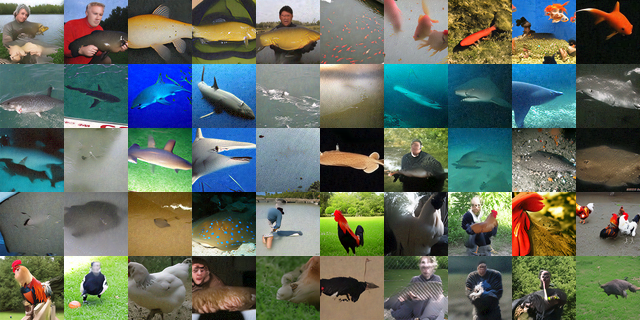}
        \caption{ImageNet64: Inversion + reconstruction, iGCT (1 NFE)}
    \end{subfigure}

    \caption{Comparison of inversion and reconstruction for CIFAR-10 (left) and ImageNet64 (right).}
    \label{fig:comparison_CIFAR-10_imagenet64}
\end{figure*}


